\documentclass[journal]{IEEEtranTIE}

\usepackage{graphicx}
\usepackage{graphics}
\usepackage{epsfig}
\usepackage{times} % assumes amsmath package installed
\usepackage{mathrsfs}
\usepackage{bm}
\usepackage{stfloats}
\usepackage{amssymb}  % assumes amsmath package installed
\usepackage{subfigure}
\usepackage[T1]{fontenc}
\usepackage[vlined,ruled]{algorithm2e}
\usepackage{booktabs}

\usepackage{float}
\usepackage{xcolor}
\usepackage{url}

\usepackage{cite}
\usepackage{picinpar}
\usepackage{amsmath}
\usepackage{flushend}
\usepackage[latin1]{inputenc}
\usepackage{colortbl}
\usepackage{soul}
\usepackage{multirow}
\usepackage{pifont}
\usepackage{color}
\usepackage{alltt}
\usepackage[hidelinks]{hyperref}
\usepackage{enumerate}
\usepackage{siunitx}
\usepackage{breakurl}
\usepackage{epstopdf}
\usepackage{pbox}

\makeatother
\usepackage{hyperref}
\usepackage{algpseudocode}
\usepackage[export]{adjustbox}
\usepackage{balance}

\soulregister{\cite}7
\soulregister{\ref}7
\begin{document}
\title{Autonomous Exploration Method for Fast Unknown Environment Mapping by Using UAV Equipped with Limited FOV Sensor}

\author{
	\vskip 1em
	
	Yinghao Zhao,
	Li Yan, \emph{Member, IEEE},
	Hong Xie, Jicheng Dai, and Pengcheng Wei

	\thanks{
	
		Manuscript received Month xx, 2xxx; revised Month xx, xxxx; accepted Month x, xxxx.
		This work was supported by the National Key Research and Development Project of China (Grant No. 2020YFD1100200), and the Science and Technology Major Project of Hubei Province under Grant (Grant No. 2021AAA010).(Corresponding author: Li Yan and Hong Xie)
		
		All authors are with School of Geodesy and Geomatics, Wuhan University, Wuhan 430079, China (e-mail: zhaoyinghao@whu.edu.cn; lyan@sgg.whu.edu.cn; hxie@sgg.whu.edu.cn; daijicheng@whu.edu.cn; wei.pc@whu.edu.cn).
	}
}
\maketitle
\begin{abstract}
Autonomous exploration is one of the important parts to achieve the fast autonomous mapping and target search. However, most of the existing methods are facing low-efficiency problems caused by low-quality trajectory or back-and-forth maneuvers. To improve the exploration efficiency in unknown environments, a fast autonomous exploration planner (FAEP) is proposed in this paper. Different from existing methods, we firstly design a novel frontiers exploration sequence generation method to obtain a more reasonable exploration path, which considers not only the flight-level but frontier-level
factors in the asymmetric traveling salesman problem (ATSP). Then, according to the exploration sequence and the distribution of frontiers, an adaptive yaw planning
method is proposed to cover more frontiers by yaw change during an exploration journey. In addition, to increase the speed and fluency of flight, a dynamic replanning strategy
is also adopted. We present sufficient comparison and evaluation experiments in simulation environments. Experimental results show the proposed exploration planner has better performance in terms of flight time and flight distance compared to typical and state-of-the-art methods. Moreover, the effectiveness of the proposed method is further evaluated in real-world environments.
\end{abstract}

%\markboth{IEEE TRANSACTIONS ON INDUSTRIAL ELECTRONICS}%
{}

\definecolor{limegreen}{rgb}{0.2, 0.8, 0.2}
\definecolor{forestgreen}{rgb}{0.13, 0.55, 0.13}
\definecolor{greenhtml}{rgb}{0.0, 0.5, 0.0}

\section{Introduction}

\IEEEPARstart{I}{ntelligent} robot has always been the goal pursued in lots of fields, and many application cases have emerged through the efforts of many researchers \cite{zhang2021novel, wang2021near}. Unmanned aerial vehicles (UAVs) as a data collection and material transportation platform with its unique advantages have been widely used in the field of surveying and mapping \cite{zheng2018multi}, environmental protection \cite{chen2020sloam}, rescue \cite{erdelj2017help}, military \cite{ramesh2020comparative}, and other fields \cite{jiang2020dvio} in recent years. However, in most operation scenarios, it is still in the state of human operation, and autonomous operation ability still lacks. As one of the key parts of UAV autonomous mapping capability, autonomous exploration has attracted extensive attention and many excellent autonomous exploration algorithms have emerged \cite{bircher2016receding, meng2017two}.
\vspace{-2mm}
\subsection{Existing Problems}
\label{EP}
Although existing exploration methods can explore environments by using frontiers or sampling viewpoints, there are still many problems to be solved. The methods of sampling viewpoints can easily generate the candidate goals, but it always causes a low exploration rate and efficiency \cite{bircher2016receding}. And most of them are using the greedy strategy, which pays attention to the local information gain but ignores the global efficiency. The work in \cite{selin2019efficient} improves the efficiency by using a frontier-based strategy to guide the "next-best-view" planning. However, they do not consider the dynamics of UAV, which will cause unsmooth exploration trajectory, low-speed flight, and lots of stop-and-go maneuvers. Although the methods \cite{meng2017two, cieslewski2017rapid} using frontiers can quickly explore the whole environment by searching frontiers and generating an exploration sequence, the process of finding and describing frontiers is always computationally expensive. FUEL \cite{zhou2021fuel} is a state-of-the-art fast and autonomous exploration algorithm. Its heuristic framework can achieve rapid and efficient UAV exploration through the designed frontier information structures (FISs) and hierarchical planning. And it can generate a smooth and high-speed exploration trajectory in high frequency. However, although this algorithm has greatly improved the exploration rate and exploration efficiency compared with other algorithms, it still faces problems affecting its exploration efficiency, such as back-and-forth maneuvers during the exploration process.
\vspace{-2mm}
\subsection{Contributions}
To reduce the back-and-forth maneuvers that cause inefficient exploration mentioned in Sect. \ref{EP}, based on the framework of FUEL, this paper proposes a novel fast autonomous exploration planner (FAEP). To improve the rationality of the exploration trajectory and the efficiency of exploration, we design a comprehensive exploration sequence generation method for global tour planning, which not only considers the flight-level factors but also innovatively considers the frontier-level factors (spatial features of frontiers) to reduce the back-and-forth exploration maneuvers. According to the spatial features, we evaluate the potential possibility of the frontier causing back-and-forth maneuvers, and get a more reasonable exploration sequence based on the possibility. After determining the local exploration target, an adaptive yaw planning strategy are designed to achieve efficient exploration by yaw change during flight, which contains two yaw planning modes: normal planning mode and two-stage planning mode. The appropriate mode can be selected to cover more areas according to the distribution of frontiers and its own motion state. In addition, to improve the stability and speed of the flight, a dynamic replanning is also adopted.

We compare our method with three typical and state-of-the-art methods in different simulation environments. The experimental results show that our method achieves a better performance than the other three methods. Compared to the two typical methods NBVP and Aeplanner, the exploration process of our method is 3-7 times faster. Compared to the state-of-the-art method FUEL, our method reduces the flight time and flight distance by more than 20$\%$. In addition, we also verify the effectiveness of our method through onboard real-world exploration. The contributions of this paper are summarized as follows:

\begin{enumerate}[1)]
	\item A comprehensive frontier exploration sequence generation method, which considers not only flight-level but frontier-level factors to generate a more reasonable global path. We innovatively consider the spatial location of the frontier and its potential to cause the low-efficiency maneuvers. The space that may cause back-and-forth maneuvers will be prioritized for exploration to improve the efficiency.
	\item An adaptive yaw planning strategy, which consists of two yaw planning modes: normal planning mode and two-stage planning mode. A smooth yaw path will be generated by different mode to cover frontiers when flying to the local target. Specifically, the two-stage planning mode will be executed to cover more areas or special small areas by yaw change at a reasonable time.  
	\item Sufficient quantitative comparison experiments are conducted in simulation. And real-world experiments are also carried out to validate our method in various environments. More details about the experiments are provided in \url{https://youtu.be/0Y671mEwJ\_A}.
\end{enumerate}

\section{RELATED WORK}
The problem of autonomous exploration has been studied by many scholars in recent years, and lots of methods from multiple angles have been published, which are mainly divided into the following three categories: sampling-based exploration, frontier-based exploration and algorithms based on machine learning which have emerged recently. This paper only discusses the previous two types of algorithms which have been widely used in various exploration tasks.
\vspace{-2mm}
\subsection{Sampling-based Exploration Methods}
Sampling-based exploration methods use randomly sampled viewpoints in the free space, which find the next best view by obtaining a path with the highest information gain\cite{connolly1985determination}. A receding horizon ``next-best-view'' planning (NBVP) is designed to explore the 3D environments by considering the information gained over the entire path in \cite{bircher2016receding}. NBVP is the first method that uses the concept of the next best view for exploration in a receding horizon fashion, and many methods are derived from this method. These methods select the path with the highest information gain in the incrementally rapidly-exploring random  trees (RRT) for UAVs to execute. The autonomous exploration planner (Aeplanner) in \cite{selin2019efficient} combined frontier exploration and NBVP to avoid getting stuck in large environments not exploring all regions, and the method also makes the process of estimating potential information gain faster by using cached points from earlier iterations. An incremental sampling and probabilistic roadmap are used in \cite{xu2021autonomous} to improve the efficiency of planning. The method \cite{respall2021fast} uses a combination of sampling and frontier-based method to reduce the impact of finding unexplored areas in large scenarios. There are also some two-stage methods \cite{cao2021tare,zhu2021dsvp} to cover the entire environment efficiently by different planning strategies in the global and local map.
\vspace{-2mm}
\subsection{Frontier-based Exploration Methods}
In contrast, the frontier-based methods are mainly comprised of two processes, finding frontiers (the boundary between known and unknown areas) and solving a sequence problem for a global path to visit frontiers. The first frontier-based exploration method is introduced by \cite{yamauchi1997frontier} to explore a generic 2D environment, which selects the closest frontier as the next goal. And then, a stochastic differential equation-based exploration algorithm\cite{shen2012stochastic} is contributed to achieve exploration in 3D environments. To achieve high-speed flight, the authors of\cite{cieslewski2017rapid} presented a method that extracts frontiers in the field of view (FOV) and selects the frontier minimizing the velocity change. For finding a reasonable frontier exploration sequence, the traveling salesman problem (TSP) is employed in \cite{meng2017two}. A wise exploration goal is selected by adopting an information-driven exploration strategy in \cite{zhong2021information}. However, many methods are facing the problems of inefficient global coverage, conservative flight trajectory, and low decision frequencies. For solving these issues,  the authors of \cite{zhou2021fuel} achieved fast exploration by adopting an incremental frontier structure and hierarchical planning. 

\section{PROPOSED APPROACH}
\label{third}
\begin{figure}[!t]
\vspace{2mm}
\centering
\includegraphics[width=\linewidth]{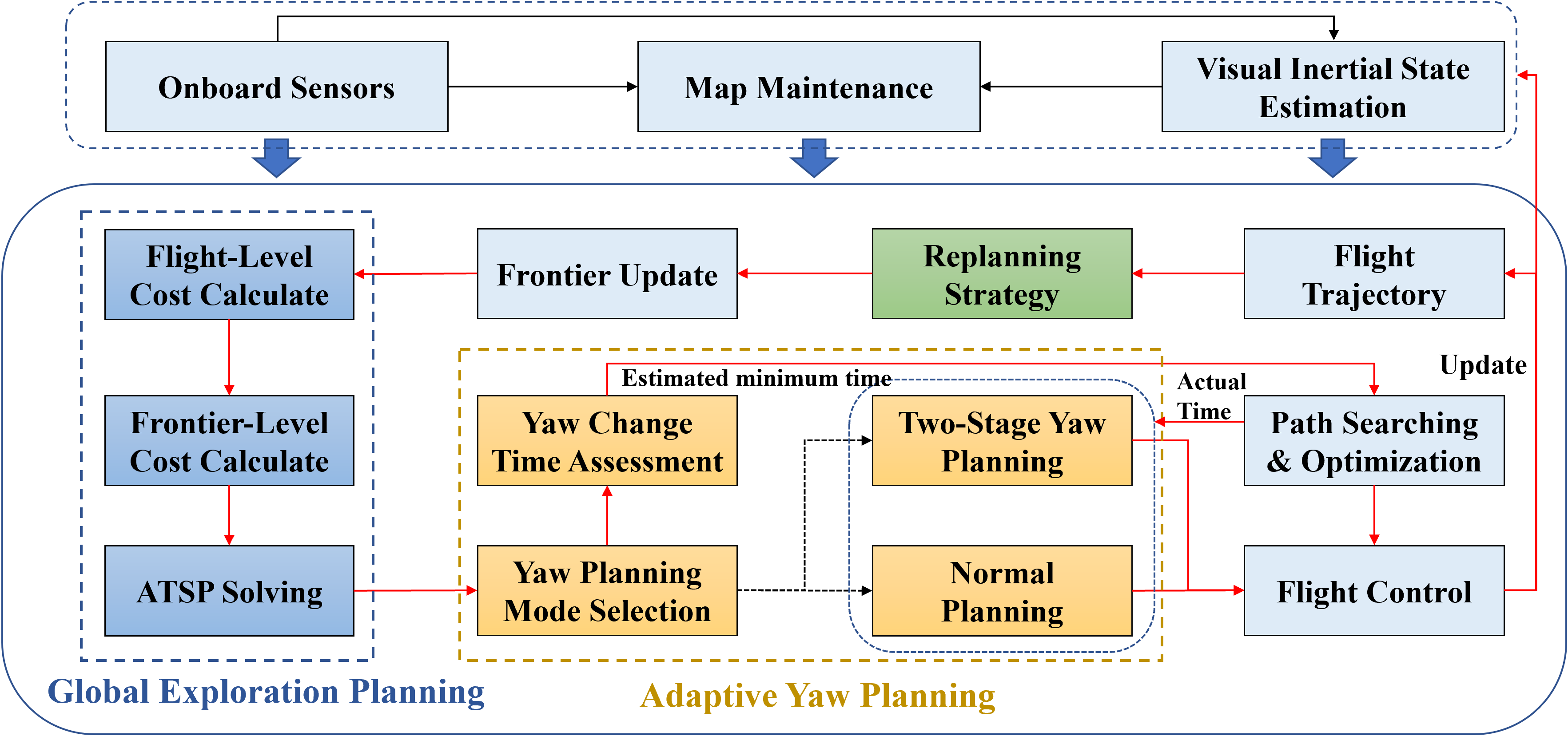}
\caption{An overview of the proposed fast autonomous exploration planner. The main flow of the planner is shown as the red line. Our main contributions are shown in dark color (blue, yellow, and green).}
\label{fig:1}
\vspace{-5mm}
\end{figure}\textbf{}
To improve the efficiency of the exploration mapping, we propose a fast autonomous exploration planner. The main operation flow of our planner is illustrated in Fig. \ref{fig:1}. First, we perform map maintenance and update frontiers based on the UAV state and the sensor data. Second, the global exploration planning is conducted by considering the flight-level and the frontier-level factors. Third, based on the result of global planning, the path planning and the adaptive yaw planning that contains two planning modes are performed to generate the high-quality local path. Finally, we will submit the planning results to the flight control for execution, and the above process will be continuously executed according to the replanning strategy.
\vspace{-2mm}
\subsection{Frontiers Exploration Sequence Generation}
\label{thirda}
The frontiers exploration sequence is crucial for the frontier-based exploration method, whose rationality directly affects the efficiency of the whole exploration process. Many methods use TSP to obtain the exploration sequence. However, most methods only take the euclidean distance between the frontiers as the cost of TSP, which is simple but insufficient. Although some methods take into account factors such as yaw change and speed direction to improve the quality of the exploration sequence on the basis of the euclidean distance, there are still some deficiencies. Different from these methods, firstly, this paper does not use the conventional TSP but uses a more reasonable Asymmetric TSP (ATSP) for the solution. Secondly, this paper not only considers flight-level factors (euclidean distance, yaw change, and speed direction change between frontiers and UAV) as the cost but also takes frontier-level factors as terms of the cost function to generate a better global exploration sequence by solving the ATSP.

\begin{figure}[t]
	\centering
	\vspace{2mm}
	\includegraphics[width= \linewidth]{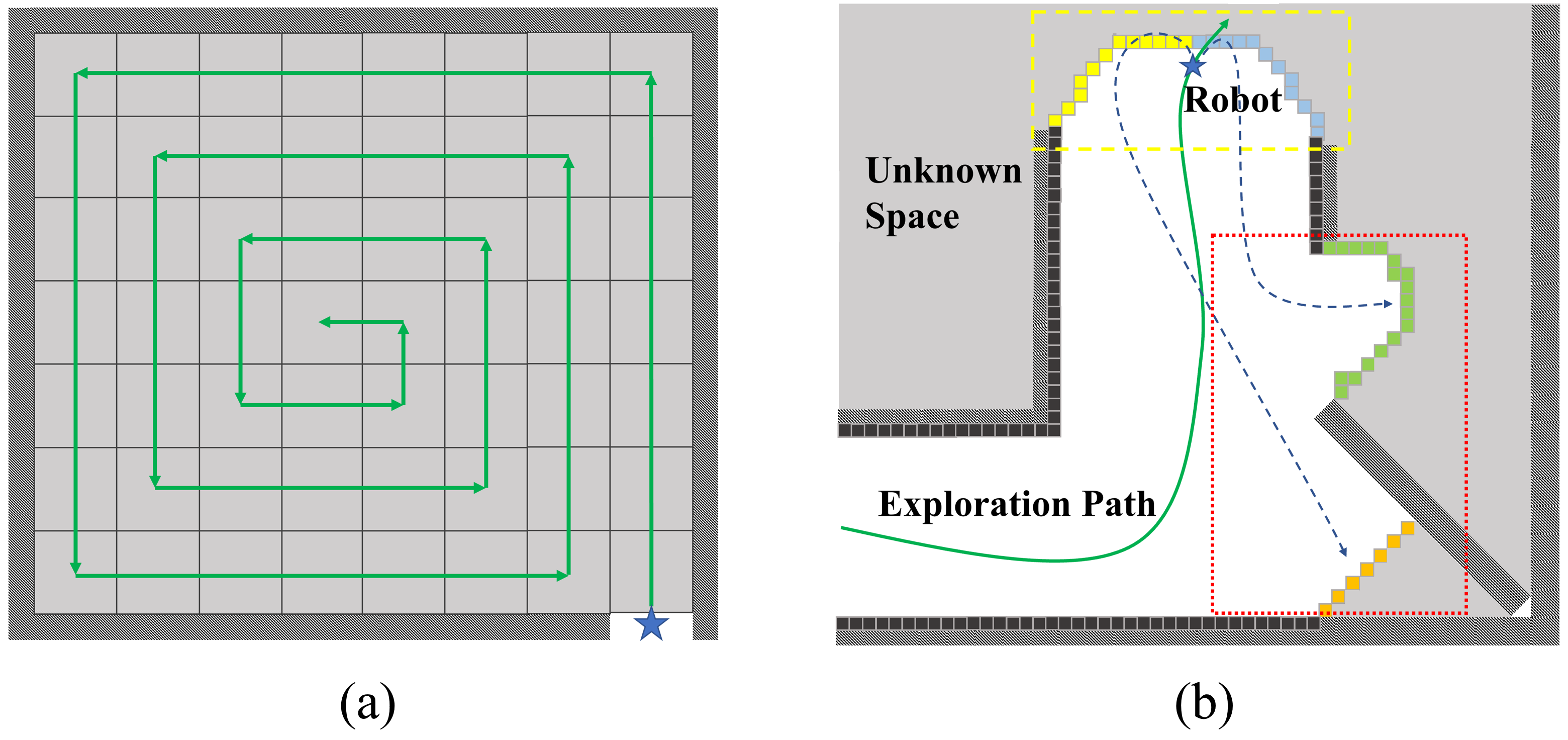}
	\vspace{-6mm}
	\caption{The subgraph (a) is a diagram of exploration process by using spiral filling algorithm. The subgraph (b) represents the exploration process without independent small area priority. The yellow box in (b) shows the frontiers being explored. The red box in (b) shows the frontiers that will cause back-and-forth path (the dotted line) in the future.}
	\label{fig:2}
	\vspace{-6mm}
\end{figure}

According to the evaluation criteria of complete coverage path planning (CCPP) problem \cite{galceran2013survey}, the fewer the back-and-forth maneuvers, the higher the efficiency. Therefore, when an unknown area needs to be explored quickly, one of the optimal strategy is to use spiral filling algorithms \cite{gonzalez2005bsa}. As shown in the Fig. \ref{fig:2}(a), spiral algorithm has low path overlap while satisfying high coverage. In addition, to prevent unnecessary back-and-forth maneuvers, small unknown areas should not be left behind where they have already passed. Otherwise as shown in Fig. \ref{fig:2}(b), there will be lots of areas that will cause the maneuvers. Inspired by this, this paper holds that when the frontier is close to the boundary of the exploration area or an independent small area, the corresponding exploration priority should be higher. If this area is not preferentially explored, it will cause back-and-forth maneuvers and reduce the efficiency of global exploration. Therefore, different from other algorithms that only consider the factors of the current flight-level, this method also takes the spatial features of the frontier into consideration, which obtains a more reasonable frontier exploration sequence by achieving boundary area priority and independent small area priority, where the boundary means the edge of the exploration area determined by mission.
\textbf{}

To achieve the boundary area priority, we calculate the boundary cost $c_{b}(k)$ for each frontier $F_{k}$ in FISs:
\begin{equation}
c_{b}(k)=\left\{\begin{aligned}
& \qquad \qquad d_{min }(k)&\ D_{k}<r_{s} \\
& d_{min }(k) \cdot\left(1+w_{d} \cdot\frac{D_{k}-r_{s}}{r_{s}}\right)&\ D_{k} \geq r_{s}
\end{aligned}\right.
\end{equation}
\begin{flalign}
\begin{aligned}
    d_{min }(k) = min(d_{x}^{k}, d_{y}^{k},d_{z}^{k}) \quad k \in\left\{1,2, \cdots, N_{cls}\right\}
\end{aligned}\label{equ:1}
\end{flalign}

\begin{figure}[t]
	\centering
	\vspace{2mm}
	\includegraphics[width=.8\linewidth]{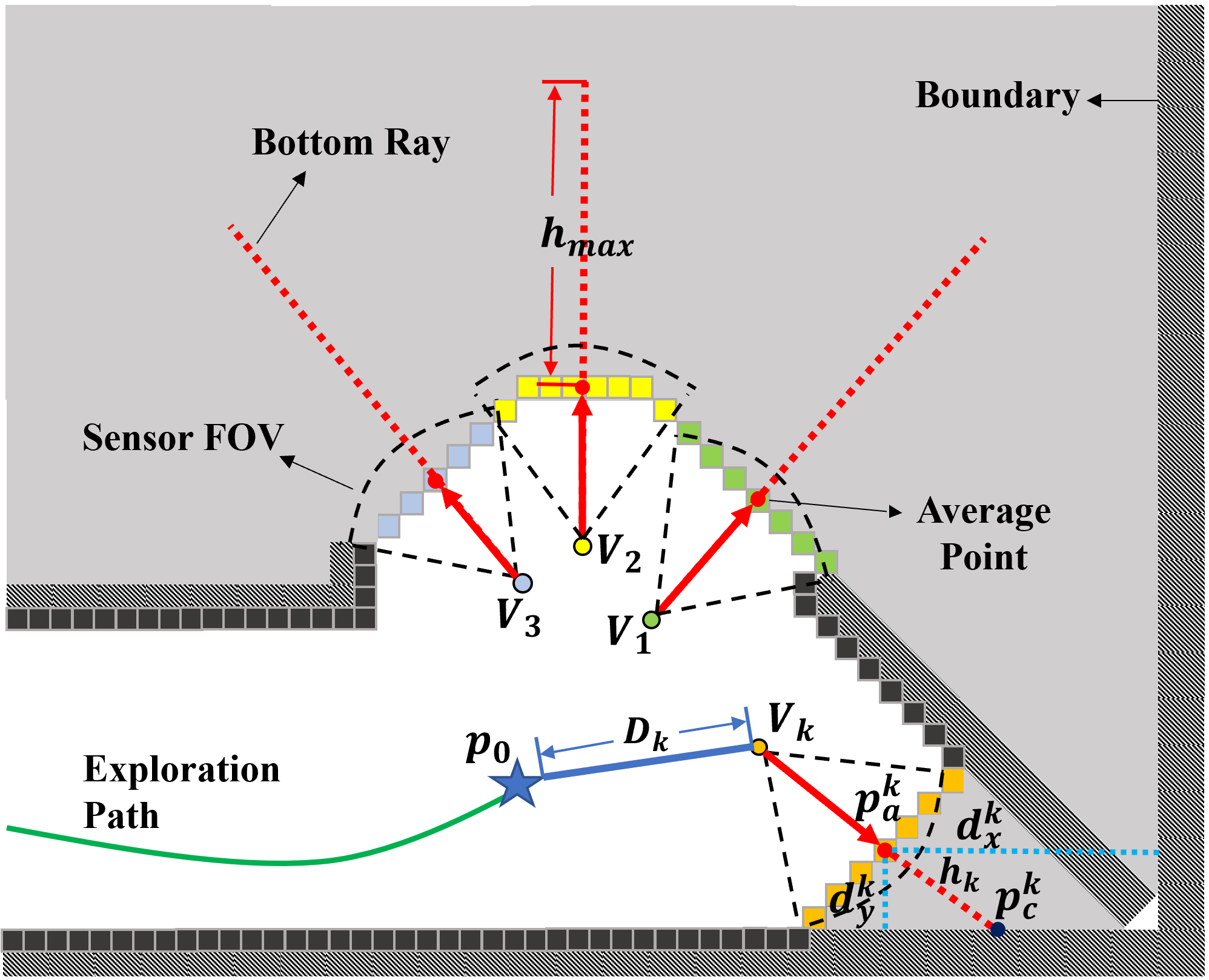}

	\caption{A diagram of evaluating the spatial features of frontiers.  $V_1$, $V_2$, $V_3$, and $V_k$ are the corresponding optimal viewpoints for covering each frontier. According to the average point $p_{a}^{k}$ of frontier $F_k$ (orange voxels), we calculate the distance (light blue line) from $F_k$ to the edge of the exploration area. And we use the Bottom Ray (red line, generated by the vector direction of $V_k$ and $p_{a}^{k}$) to roughly evaluate the scope of the unknown area (gray area) behind each frontier (color voxels). Once the ray touches the occupied, free voxel, boundary, or maximum distance $h_{max}$, we regard the distance $h_k$ of ray (red dashed line) as the scope of the frontier.}
	\label{fig:3}
	\vspace{-6mm}
\end{figure}

Where $N_{cls}$ represents the number of frontiers. As shown in Fig. \ref{fig:3}, $d_{x}^{k}$,  $d_{y}^{k}$,  $d_{z}^{k}$ is the shortest distance from the frontier to the boundary along the X, Y, Z-axis respectively, and X, Y, Z-axis are the same as the initial body frame of UAV at takeoff. $r_{s}$ is the maximum range of the sensor and $D_{k}$ is the distance between the viewpoint $V_{k}$ of the frontier $F_{k}$ and the current position $p_0$. We directly regard the minimum distance $d_{min }(k)$ as $c_{b}(k)$ when the frontier is close to the current position ($D_{k} < r_{s}$). The farther the frontier is from the boundary, the greater the boundary cost $c_{b}(k)$ is. To raise the priority of the boundary frontiers, we regard $c_{b}(k)$ as a member of the cost function in ATSP to obtain a sequence where the frontiers near the boundary will be explored in priority. Meanwhile, to avoid exploring the frontier close to the boundary but far from the current position, we penalize the frontiers by a positive penalty factor $w_{d}$ when $D_{k}$ is greater than $r_{s}$. In this way, the larger $D_{k}$ is, the larger the frontier cost is.

To achieve the independent small area priority, a method called Bottom Ray is designed as shown in Fig. \ref{fig:3}. We use the ray (red line) to roughly detect the range of the unknown area behind the frontier, and then the probability $c_{s}^{k}$ that the area is an independent small area can be evaluated according to the detection result. The main process is as follows: Firstly, we obtain the frontiers whose distance $D_{k}$ between the viewpoints $V_k (p_k,  \xi_k)$ and the current position $p_0$ of UAV is less than $D_{thr}$, where $D_{thr}$ is the distance threshold to avoid raising the priority of too far frontiers. And we use the same method in\cite{zhou2021fuel} to generate the viewpoint $V_k$, which contains a position  $p_k$ and a yaw angle $\xi_k$. The method can be divided into three steps: 1) Uniformly sampling points in the cylindrical coordinate system whose origin locates at the frontier's center. 2) The points with coverage higher than a threshold are reserved and sorted in descending order of coverage. 3) Selecting the viewpoint that needs less yaw change and flight distance as the final viewpoint. Secondly, if a frontier is adjacent to the obstacle in both directions perpendicular to the Z-axis at the same time, we believe the unknown area behind the frontier is an independent area similar to rooms or corridors, which should be explored in priority. For this case, we set $c_{s}(k)$ = 1. If not, execute the next step. Thirdly, the vector $\pmb{\overrightarrow{p_kp_{a}^{k}}}$ (red line in Fig. \ref{fig:3}) from the viewpoint $V_k$ to the average point $p_{a}^{k}$ of $F_k$ is calculated, and we extend it by mapping resolution until it touches the occupied, free voxel, boundary or exceeds the maximum distance $h_{max}$, then a bottom point $p_{c}^{k}$ and the distance $h_k$ between $p_{a}^{k}$ and $p_{c}^{k}$ (red dashed line in Fig. \ref{fig:3}) are obtained. And we calculate the independent small area probability $c_{s}(k)$ for these frontiers by:
\begin{equation}
c_{s}(k)=\frac{h_{max}-h_{k}}{h_{max}}
\end{equation}
\textbf{}
Finally, we regard $c_{s}(k)$ and $c_{b}(k)$ as the factors of frontier-level, and integrate them with flight-level factors used in \cite{zhou2021fuel} into the cost matrix $M_{tsp}$ of ATSP as follows: 
\begin{flalign}
	\begin{aligned}
		&{M}_{{tsp}}(0, k)=t_{lb}\left(V_{0}, 
		~V_{k}\right)+w_{{c}} \cdot c_{{c}}\left(V_{k}\right) \\
		&\quad+w_{{b}} \cdot c_{b}(k)-w_{{f}} \cdot c_{s}(k) \\
	\end{aligned}   \label{equ:2}
\end{flalign}
\begin{flalign}
	\begin{aligned}
		&t_{lb}\left(V_{0}, ~V_{k}\right)= \max 
		\left\{\frac{length\left(p_{0}, 
			p_{k}\right)}{v_{max }}\right. , \\
		&\left.\frac{ \min \left(\left|\xi_{0}-\xi_{k}\right|, 2 
			\pi-\left|\xi_{0}-\xi_{k}\right|\right)}{\dot{\xi}_{max }}\right\}
	\end{aligned}
\end{flalign}
\begin{flalign}
	c_{c}\left(V_{k}\right)= \cos ^{-1} \frac{\left(p_{k}-p_{0}\right) \cdot 
		v_{0}}{\left\|p_{k}-{p}_{0}\right\|\left\|v_{0}\right\|}
\end{flalign}

Where $V_0$  indicates the current state of UAV, which contains the current position $p_0$ and yaw angle $\xi_0$. $V_k$ represents the viewpoint of frontier $F_k$. $v_0$ is the current speed of UAV. $t_{lb} (V_0, V_k)$ and $c_c (V_k)$  represents flight-level factors such as distance, yaw change, and speed change. $w_{c}$, $w_{b}$, and $w_{f}$ are the corresponding weights of the three cost terms respectively. According to the Equ. \ref{equ:2}, the frontier with the minimum cost should be the frontier that has high probability of causing the back-and-forth maneuvers, and it should be the best frontier that needs to be explored promptly to reduce the low-efficiency exploration in the later stage. The calculation method of the rest of $M_{tsp}$ is as follows:
\begin{figure}[!t]
	\vspace{2mm}
	\centering
	\includegraphics[width=.9\linewidth, height=5.2cm]{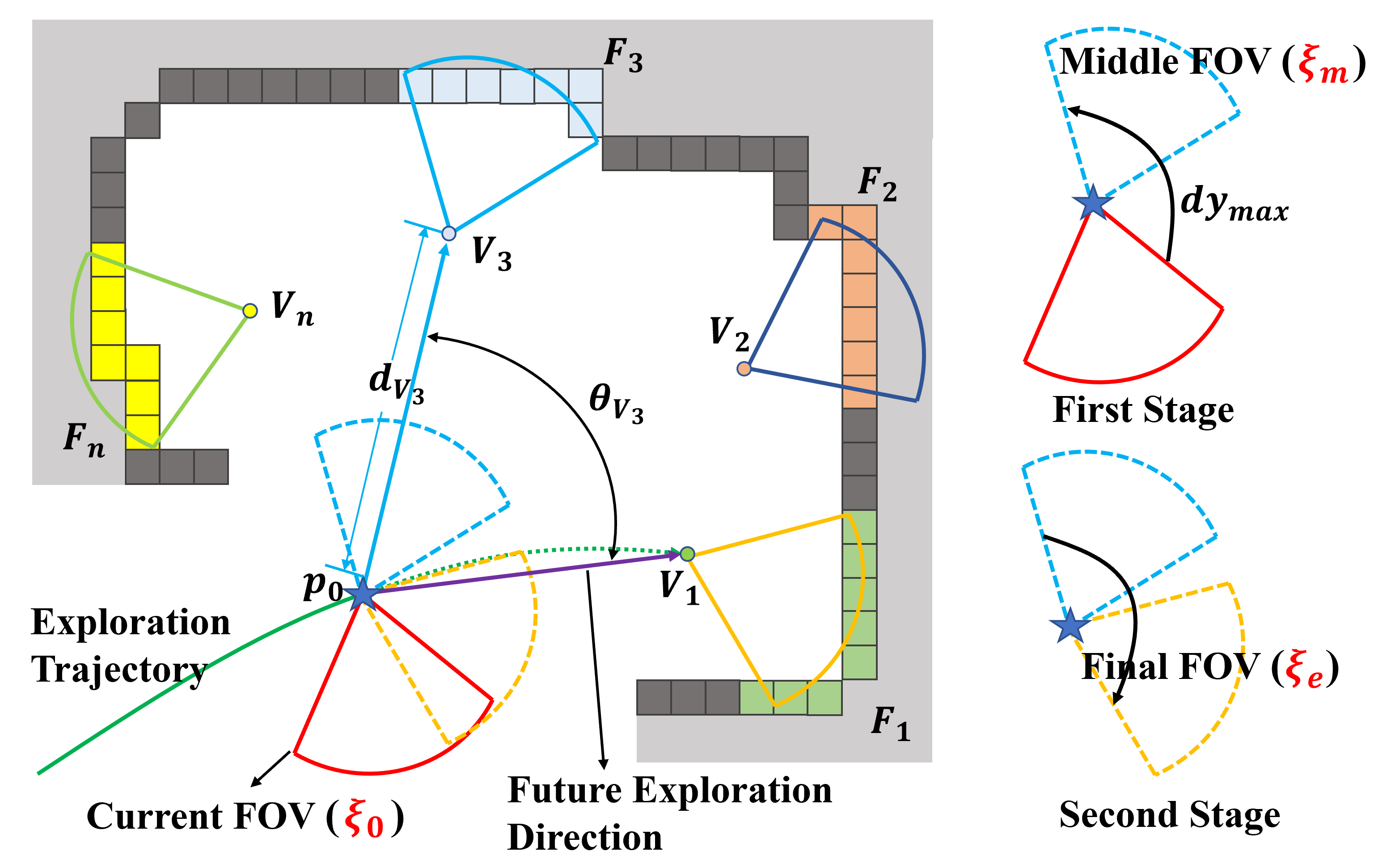}
	\caption{The schematic of the proposed two-stage yaw planning for the case of multiple viewpoints in the local range. In first stage, we calculate the middle yaw $\xi_m$ (blue FOV) according to the magnitude of the yaw change, and then the corresponding yaw planning from the current yaw $\xi_0$ (red FOV) to the middle yaw is conducted to cover more area during the flight process (green dashed line). In second stage, the yaw planning from the middle yaw to the final yaw $\xi_e$ (yellow FOV) is conducted to explore the target frontier (green grids).}
	\label{fig:4}
	\vspace{-3mm}
\end{figure}\textbf{}
\begin{flalign}
	\begin{aligned}
		&M_{tsp}\left(k_{1}, k_{2}\right)=M_{tsp}\left(k_{2}, 
		k_{1}\right) \\
		&=t_{lb}\left({V}_{k_{1}}, {V}_{k_{2}}\right), k_{1}, k_{2} 
		\in\left\{1,2, \cdots, N_{cls}\right\}
	\end{aligned}
\end{flalign}
\begin{flalign}
	M_{tsp}(k, 0)=0, k \in\left\{0,1,2, \cdots, N_{cls}\right\}
\end{flalign}

\subsection{Adaptive Yaw Planning}
\label{thirdb}
When the UAV is equipped with limited FOV sensors, yaw planning becomes another important thing besides path planning. Through a large number of experiments, we observe that when UAV goes to the local target, there are often multiple areas that can be covered by yaw change without affecting the flight direction. Once there is an excellent yaw planning, the UAV can explore more areas during one planning task, and the efficiency of exploration will be improved.

However, covering more areas often needs larger yaw change, and means more time cost. If the planner keeps using this strategy during the exploration, it will cause the exploration process at an extremely slow speed. Therefore, the timing of using this strategy is also important. Based on this, this paper designs an adaptive yaw planning strategy. Different from other yaw planning methods, two planning modes are adopted: normal yaw planning and two-stage yaw planning. The normal yaw planning is used to plan a smooth yaw trajectory based on the current yaw of robot and the viewpoint of the target frontier. The two-stage yaw planning is designed to utilize a larger range of yaw change to cover more frontiers when the time required for two-stage change is less than or approximately equal to the flight time of the current trajectory. The main process is described in adaptive yaw planning strategy, where $VPs$ is a set containing all viewpoints. $V_{1}$ and $X_{0}$ are the next target viewpoint and current motion state respectively.

At first, due to the limited range of the sensor, only the area whose distance from the current position $p_{0}$ is less than sensor range $r_{s}$ can be covered. Therefore, we use \textbf{ViewpointsInLocal()} to calculate the number of viewpoints $V_{k}$ whose distance $D_{k}$ between $V_{k}$ and current position $p_{0}$ is less than the threshold $d_{thr}$ ($d_{thr}<r_{s}$) and the two points are intervisible. In addition, the angle ${\theta}_{V_{k}}$ between $\pmb{\overrightarrow{p_0V_{k}}}$ and $\pmb{\overrightarrow{p_0V_{1}}}$ need to be less than 90 degrees to ensure the ability of coverage (Line 1). And then, if there are viewpoints that meet the above conditions, we adopt the two-stage yaw planning mode (Line 2-11). Otherwise, the normal yaw planning method is used (Line 12-15). If the two-stage planning is adopted, as shown in Fig. \ref{fig:4}, we use \textbf{FindMiddleYaw()} to calculate the yaw change between each viewpoint and the current state, and find the viewpoint (blue FOV) with the largest yaw change $\xi_m$ that meets the time condition (Line 3). Later, according to the geometric relationship between $\xi_m$, the current yaw $\xi_0$, the yaw $\xi_e$ of the target frontier, and the max angular velocity $\dot{\xi}_{\max }$, the minimum time $T_{min}$ required for the two yaw changes is roughly calculated by \textbf{EstimateMinYawTime()}: 
\begin{flalign}
	T_{1}=\frac{\min \left(\left|\xi_{m}-\xi_{0}\right|, 2 
		\pi-\left|\xi_{m}-\xi_{0}\right|\right)}{\dot{\xi}_{\max }}
\end{flalign}
\begin{flalign}
	T_{2}=\frac{\min \left(\left|\xi_{e}-\xi_{m}\right|, 2 
		\pi-\left|\xi_{e}-\xi_{m}\right|\right)}{\dot{\xi}_{\max }}
\end{flalign}
\begin{flalign}
	T_{min}=\tau \cdot\left(T_{1}+T_{2}\right)
\end{flalign}

Where $\tau$ is expansion coefficient that can be used to improve the feasibility of yaw change. To avoid excessive influence of yaw change on flight speed, we only carry out two-stage yaw planning when the time of yaw change is less than the flight time (Line 6). However, once the area is probably a small area judged by Sect. \ref{thirda}, the two-stage yaw planning will be used to cover it without hesitation for avoiding possible back-and-forth maneuvers later (Line 6). And then, if the above conditions are met, the $T_{min}$, current motion state $X_0$ and the position $p_e$ of the next target viewpoint will be provided for \textbf{PathPlanningAndOpt()} introduced in \ref{sub:pathsearch} to generate a flight path (Line 7), where $T_{min}$ is regarded as the minimum flight time constraint. Finally, if the actual flight time $T_{real}$ generated by \textbf{PathPlanningAndOpt()} is more than $T_{min}$, the two stage yaw planning will be executed by \textbf{YawPlanning()} (Line 8-11, 12-15). In this function, we use an uniform B-spline to represent the trajectory of yaw angle $\phi(t)$, which is parameterized by the  N+1 control points $\Phi:=\{\phi_0, ... \phi_n\}$ and knot span $\delta t_{\phi}$. $T$ is the total time of the trajectory. Due to the convex hull property of B-spline, the smoothness and dynamic feasibility of the trajectory can be optimized by changing the control points $\Phi$ and solving the problem:
\begin{flalign}
	\begin{aligned}
		\underset{\xi_{c p}}{\arg \min } \  \gamma_{1} f_{s}+\gamma_{2} 
		&\left(\phi\left(t_{0}\right)-\xi_{0}\right)+\gamma_{3}\left(\phi(T)-\xi_{e}\right) \\
		&+\gamma_{4}\left(f_{\dot{\xi}}+f_{\ddot{\xi}}\right)
	\end{aligned}
\end{flalign}

Where $f_s$ represents smoothness. The second and third terms are soft waypoint constraint enforcing $\phi(t)$ to pass through current yaw $\xi_0$ and target yaw $\xi_e$. The last two terms are the soft constraints for the dynamic feasibility of angular velocity and acceleration. $\gamma_{1}$, $\gamma_{2}$, $\gamma_{3}$, and $\gamma_{4}$ are the corresponding weights of each item. The calculation methods of $f_s$, $f_{\dot \xi}$, and $f_{\ddot \xi}$ are similar to  \cite{zhou2021raptor}. 

\begin{table*}[!ht]
	\begin{center}
		\caption{EXPLORATION STATISTIC IN THE THREE SCENARIOS}
		\label{tab:1}
		\centering
		\setlength{\tabcolsep}{2.7mm}{
			\renewcommand{\arraystretch}{1.1} {% Default value: 1
				\begin{tabular}{cccccccccccccc}
					\toprule
					\multirow{2}{*}{\textbf{Scene}} & \multirow{2}{*}{\textbf{Method}} & \multicolumn{4}{c}{\textbf{Exploration time (s)}}                                                                                   & \multicolumn{4}{c}{\textbf{Flight distance (m)}}                                                                                     & \multicolumn{4}{c}{\textbf{Coverage (m$^3$)}}                                                                                             \\ \cline{3-14} 
                                &                                  & \textbf{Avg}   & \textbf{Std} & \textbf{Max}   & \textbf{Min}   & \textbf{Avg}  & \textbf{Std} & \textbf{Max}   & \textbf{Min}   & \textbf{Avg}    & \textbf{Std} & \textbf{Max}   & \textbf{Min}   \\
\multirow{4}{*}{Maze1}          & Aeplanner                        & 328.1                           & 43.7                          & 400.9                           & 222.07                          & 199.0                           & 26.3                           & 249.1                           & \textbf{140.9}& 859.7                            & 24.0                          & 899.0                            & 805.3                            \\
                                & NBVP                             & 650.3                           & 139.8                         & 928.9                           & 418.2                           & 302.7                           & 60.4                           & 413.3                           & 198.6                           & 859.8                            & 41.5                          & \textbf{950.3} & 756.8                            \\
                                & FUEL                             & 158.9                           & 8.2                           & 175.6                           & 146.6                           & 219.0                           & 10.5                           & 242.4                           & 200.6                           & \textbf{901.8} & 3.2                           & 909.0                            & \textbf{896.0} \\
                                & Proposed                         & \textbf{116.3}& \textbf{6.4}& \textbf{128.3}& \textbf{103.1}& \textbf{164.6}& \textbf{10.0}& \textbf{182.9}& 141.6                           & 892.9                            & \textbf{2.3}& 898.4                            & 889.2                            \\
                                \hline
\multirow{4}{*}{Maze2}      & Aeplanner                        & 302.0                           & 31.0                          & 346.9                           & 235.0                           & 171.7                           & 17.1                           & 198.2                           & 126.7                           & 726.8                            & 14.3                          & 745.4                            & 680.2                            \\
                                & NBVP                             & 517.9                           & 76.3                          & 653.7                           & 336.2                           & 248.7                           & 37.8                           & 320.8                           & 159.6                           & 733.3                            & 23.5                          & \textbf{768.9} & 689.0                            \\
                                & FUEL                             & 132.4                           & 7.6                           & 145.9                           & 117.5                           & 179.7                           & 10.8                           & 196.2                           & 150.5                           & \textbf{753.3} & 3.2                           & 762.7                            & \textbf{748.2} \\
                                 & Proposed                         & \textbf{105.5}& \textbf{6.0}& \textbf{116.0}& \textbf{95.5} & \textbf{141.1}& \textbf{10.0}& \textbf{159.1}& \textbf{122.2}& 747.1                            & \textbf{2.1}& 751.9                            & 742.8                            \\
                                \hline
\multirow{4}{*}{Outdoor}     & Aeplanner                        & 661.2                           & 101.0                         & 856.2                           & 463.8                           & 405.2                           & 52.5                           & 505.0                           & 308.3                           & 1763.5                           & 9.7                           & \textbf{1783.7}& 1746.1                           \\
                                & NBVP                             & 917.1                           & 185.9                         & 1390.4                          & 630.5                           & 429.9                           & 76.4                           & 591.6                           & 300.8                           & 1683.3                           & 76.5                          & 1777.3                           & 1501.9                           \\ 
                                 & FUEL                             & 177.0                           & 9.5                           & 201.7                           & 161.1                           & 260.1                           & 14.4                           & 298.5                           & 236.4                           & \textbf{1773.9}& \textbf{4.3}& 1781.7                           & \textbf{1767.6}\\
                                & Proposed                         & \textbf{139.7}& \textbf{8.5}& \textbf{154.8}& \textbf{121.6}& \textbf{208.2}& \textbf{11.7}& \textbf{232.0}& \textbf{186.4}& 1755.9                           & 10.5                          & 1776.9                           & 1735.0                             \\
                                \bottomrule
				\end{tabular}
		}}
	\end{center}
	\vspace{-6mm}
\end{table*}

\begin{figure*}[!ht]
    \subfigure[]{
    \begin{minipage}[t]{0.31\linewidth}
    \centering
    \includegraphics[width=2.2in]{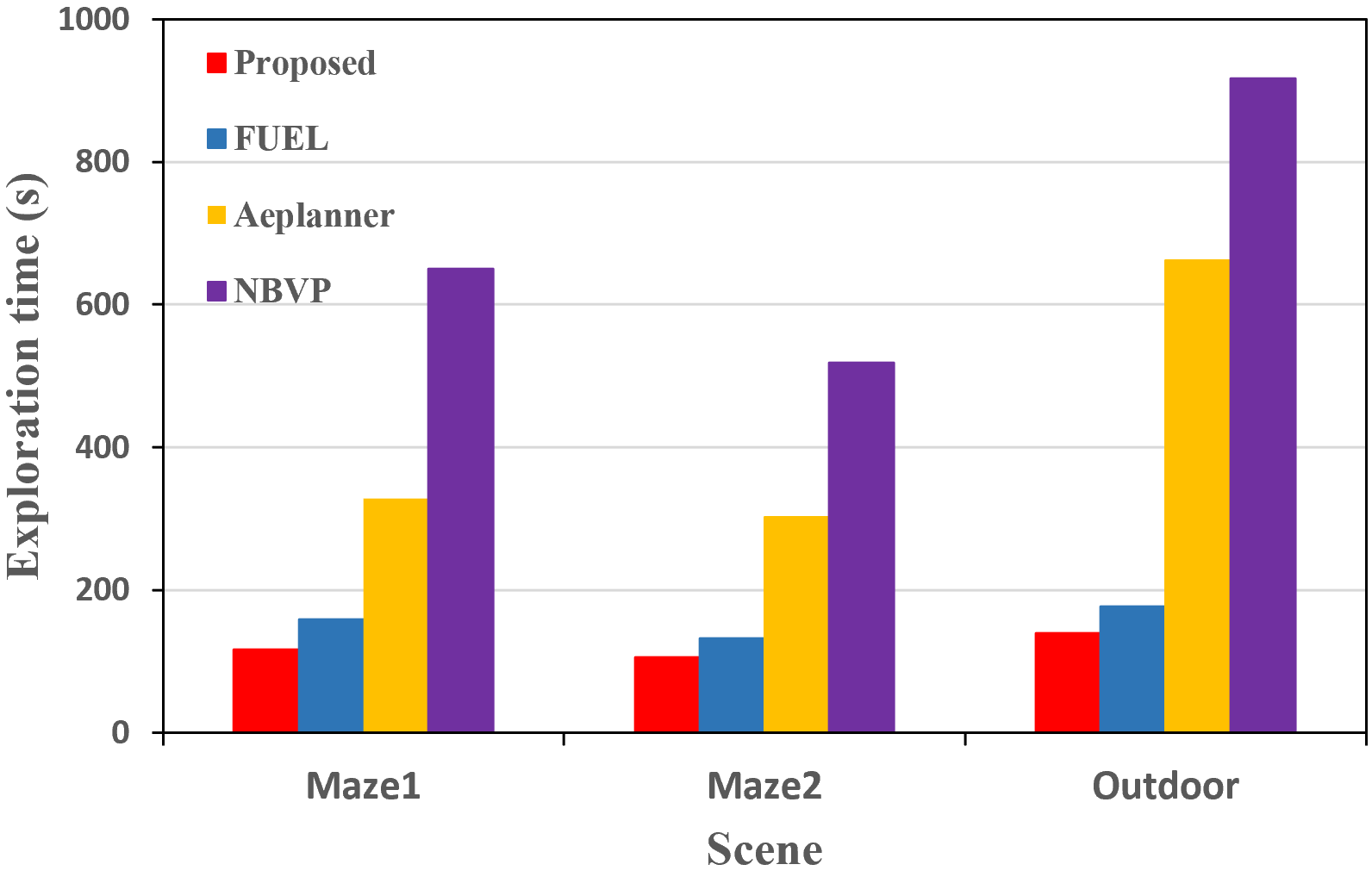}
    \end{minipage}
    }
    \subfigure[]{
    \begin{minipage}[t]{0.31\linewidth}
    \centering
    \includegraphics[width=2.2in]{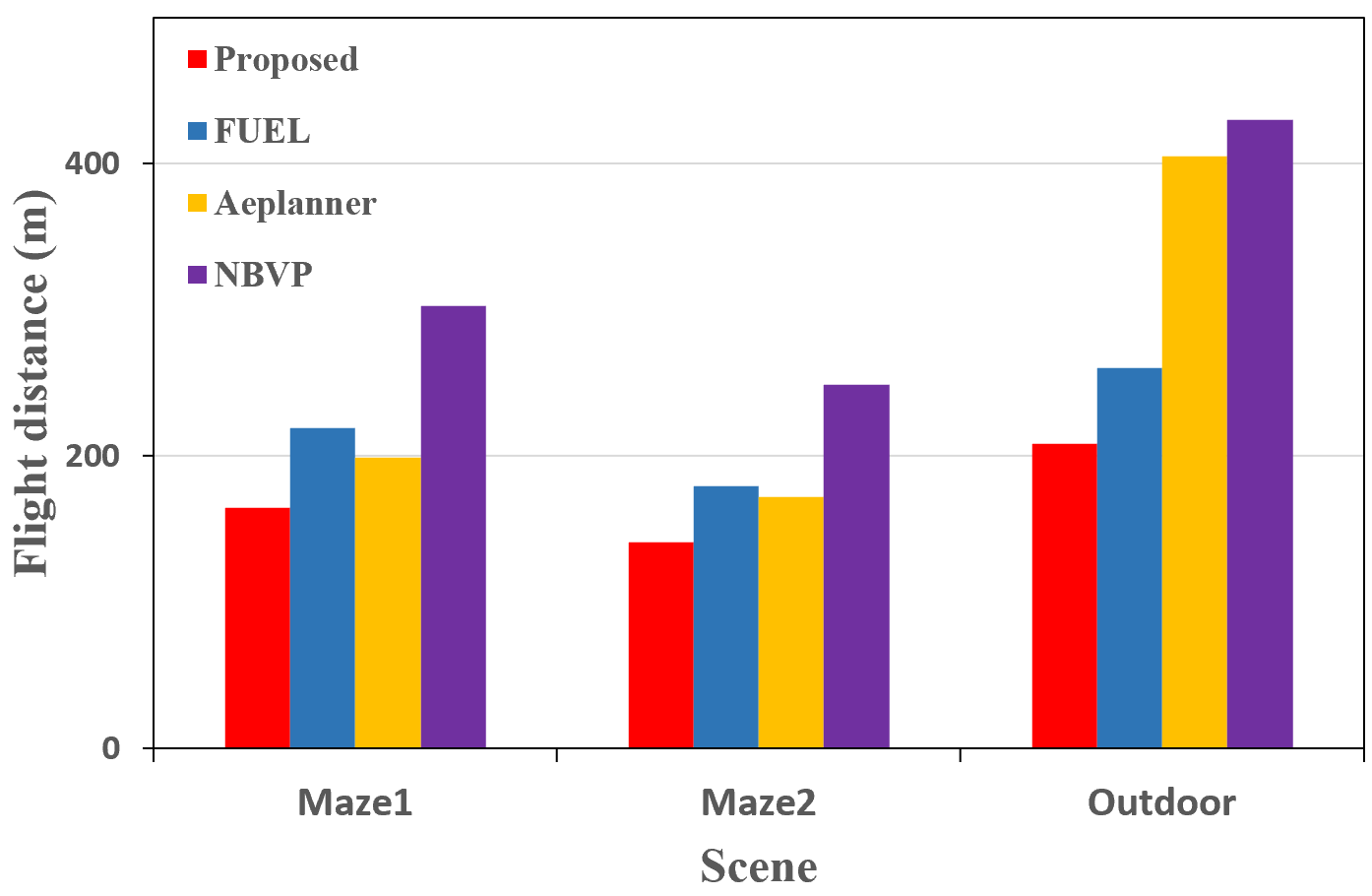}
    \end{minipage}
    }
    \subfigure[]{
    \begin{minipage}[t]{0.31\linewidth}
    \centering
    \includegraphics[width=2.2in]{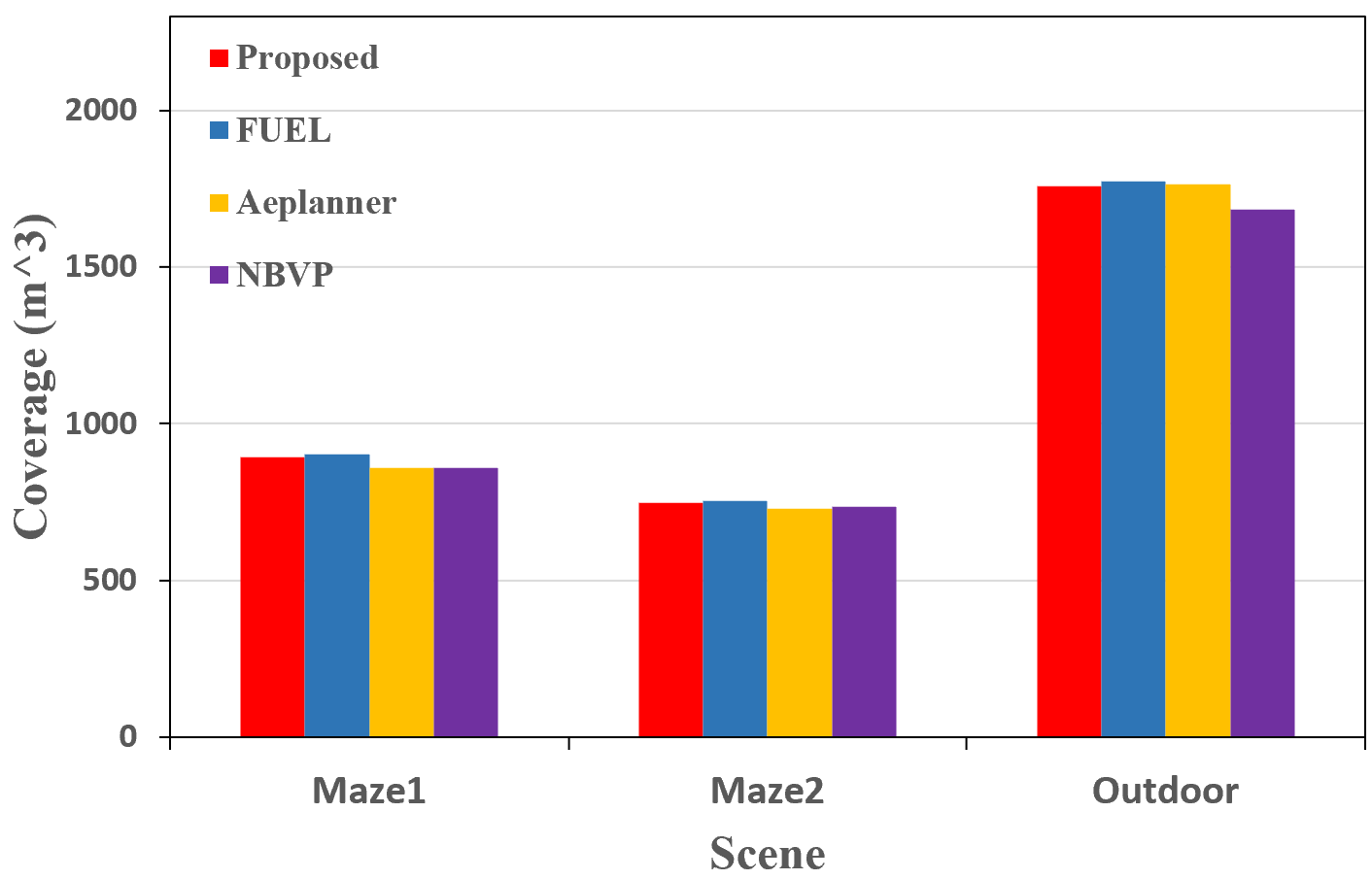}
    \end{minipage}
    }
    \centering
    \caption{(a), (b), and (c) are average exploration time, average flight distance, and average coverage presented in Table. \ref{tab:1} respectively. }
    \label{fig:table}
    \vspace{-5mm}
\end{figure*}

\floatname{algorithm}{Algorithm}
\renewcommand{\algorithmicrequire}{\textbf{Input:}}
\renewcommand{\algorithmicensure}{\textbf{Output:}}  

\begin{algorithm}[!t]
    \caption{Adaptive Yaw Planning Strategy}
    \begin{algorithmic}[1]
        \Require $VPs, V_1 (p_e,\xi_{e}), X_0 (V_0,v_0,a_0), D_{k}, c_{s}(k)$
        \Ensure yaw Trajectory $Y$ 
        \State $N_{v} \gets \textbf{ViewpointsInLocal}(VPs)$
        \State \textbf{if} $N_{v} > 1$ \textbf{then}
	    \State $\quad \xi_{m} \gets \textbf{FindMiddleYaw}(N_{v})$  
	    \State $\quad T_{1},T_{2} \gets \textbf{EstimateMinYawTime}(\xi_{0},\xi_{m},\xi_{e})$   
	    \State $\quad T_{min} \gets \tau \cdot(T_{1}+T_{2}), R \gets T_{1}/T_{min}$
	    \State \quad \textbf{if} $T_{min} <= D_{k}/v_0 \ ||\  c_{s}(k) > 0.5$ \textbf{then}
	    \State $\qquad T_{real} \gets \textbf{PathPlanningAndOpt}(X_0,p_e,T_{min})$ 
        \State \qquad \textbf{if} $T_{real} >= T_{min}$ \textbf{then}
	    \State $\qquad \quad Y_1 \gets \textbf{YawPlanning}(\xi_{0},\xi_{m},T_{real}*R)$  
	    \State $\qquad \quad Y_2 \gets \textbf{YawPlanning}(\xi_{m},\xi_{e},T_{real}*(1-R))$  
	    \State \qquad \quad \Return{$Y(Y_{1}, Y_{2})$} 
        \State $T_{min} \gets \textbf{EstimateMinYawTime}(\xi_{0},\xi_{e})$  
        \State $T_{real} \gets \textbf{PathPlanningAndOpt}(X_{0},p_{e},T_{min})$  
        \State $Y \gets \textbf{YawPlanning}(\xi_{0},\xi_{e},T_{real})$  
        \State \Return{$Y$}  
    \end{algorithmic} 
\end{algorithm}
\vspace{-1mm}
\subsection{Path Searching and Optimization}\label{sub:pathsearch}
Another core component of achieving fast exploration is path planning, which is used to generate a safe, smooth, and dynamically feasible trajectory for fast flight. According to the content, path planning can be divided into two parts: path searching and path optimization. In path searching, kinodynamic path searching \cite{zhou2019robust} is widely adopted, which can use motion primitives respecting the UAV dynamic to find a collision-free and dynamic feasible path. However, when UAV is in special scenes, such as searching path from inside to outside in a house or passing through a narrow space, if only the conventional kinodynamic path searching is adopted, the search process will take a relatively long time or even fail due to the discrete control space. To reduce the occurrence of this situation, we adopt our previous work \cite{zhao2021robust} to search the initial path, which use a guiding path to improve the stability of kinodynamic path searching. In path optimization, we optimize the initial path by applying the B-spline trajectory optimization \cite{zhou2019robust}. The method also utilizes the convex hull property of uniform B-splines to generate smooth, safe, and dynamically feasible trajectory by minimizing the cost function.
\vspace{-3mm}
\subsection{Adaptive Dynamic Replanning}
The speed of the target point is usually set to zero by default, and the cost time of each replanning is dynamic and unknown. Therefore, a low-frequency replanning strategy will cause low speed exploration due to the stop-and-go maneuvers. Besides, if the current position is used as the starting point for planning, the flight will be unstable when planning process takes a long time due to a long distance between the starting point of the new path and the current position of UAV. To solve these problems, this paper adopts adaptive dynamic starting point for exploration replanning inspired by \cite{tordesillas2019faster}.  In the i-th planning step, we do not use the current position as the starting point of the planning but select the position at the time $t_i$ in the future as the starting point of the planning, and $t_i$ is not constant but determined by the previous planning result:
\begin{flalign}
	t_{i}= \max \left(\rho \cdot t_{i-1}, t_{min }\right)
	\label{equ:15}
\end{flalign}

Where $t_i, t_{i-1}$ represents the cost time of i-th and i-1-th planning step respectively.  $t_{min}$  is the minimum time for one planning. $\rho$ is the expansion coefficient to improve reliability. If the planning is successful and the actual planning time is less than $t_i$, update the path after time $t_i$ with the planning result. Otherwise, execute replanning. Besides, to maintain the speed and fluency of the flight, we make a replanning when the duration of the remaining flight path is less than 1s.
\section{Experimental Results}

\begin{figure*}[!ht]
    \subfigure[]{
    \begin{minipage}[t]{0.31\linewidth}
    \centering
    \label{scene1}
    \includegraphics[width=2.2in, height=1.2in]{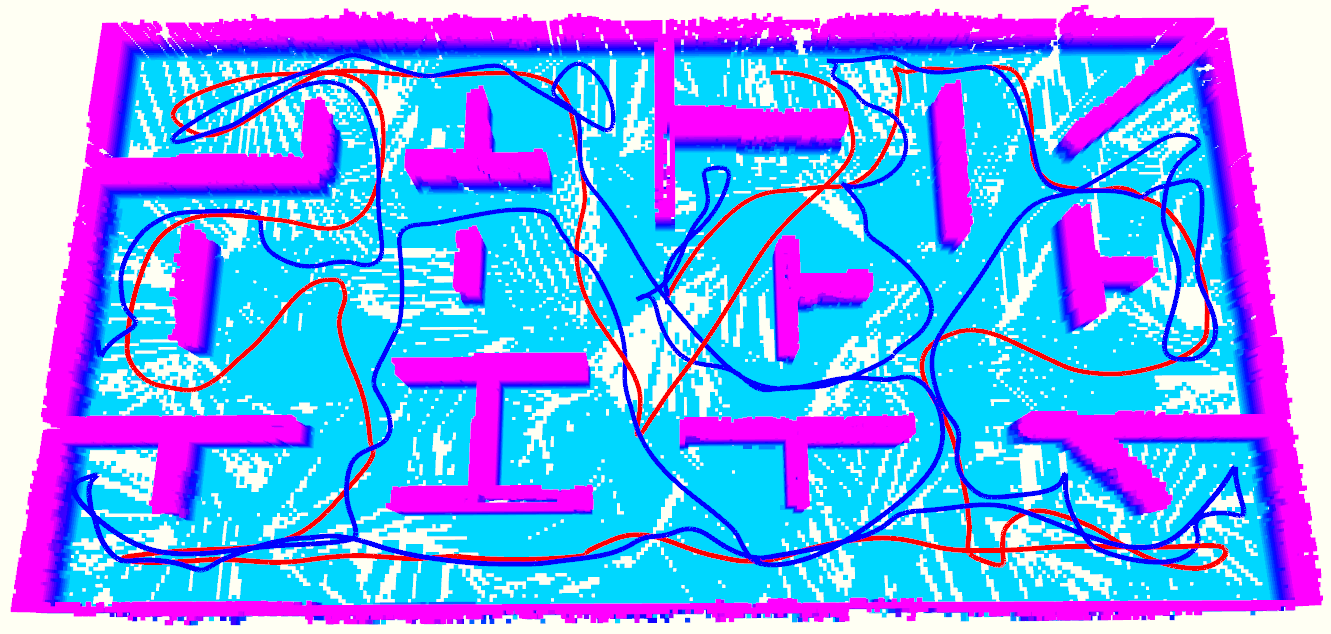}
    \end{minipage}
    }
    \subfigure[]{
    \begin{minipage}[t]{0.31\linewidth}
    \centering
    \label{scene2}
    \includegraphics[width=1.4in]{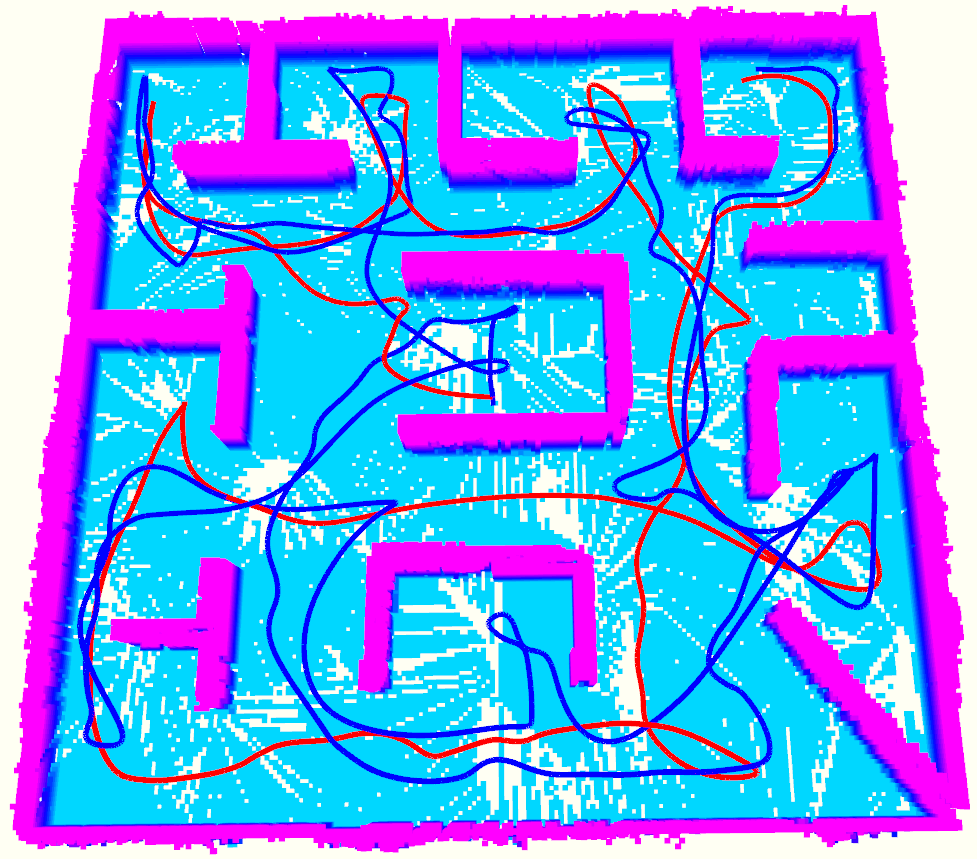}
    \end{minipage}
    }
    \subfigure[]{
    \begin{minipage}[t]{0.31\linewidth}
    \centering
    \label{scene3}
    \includegraphics[width=2.2in]{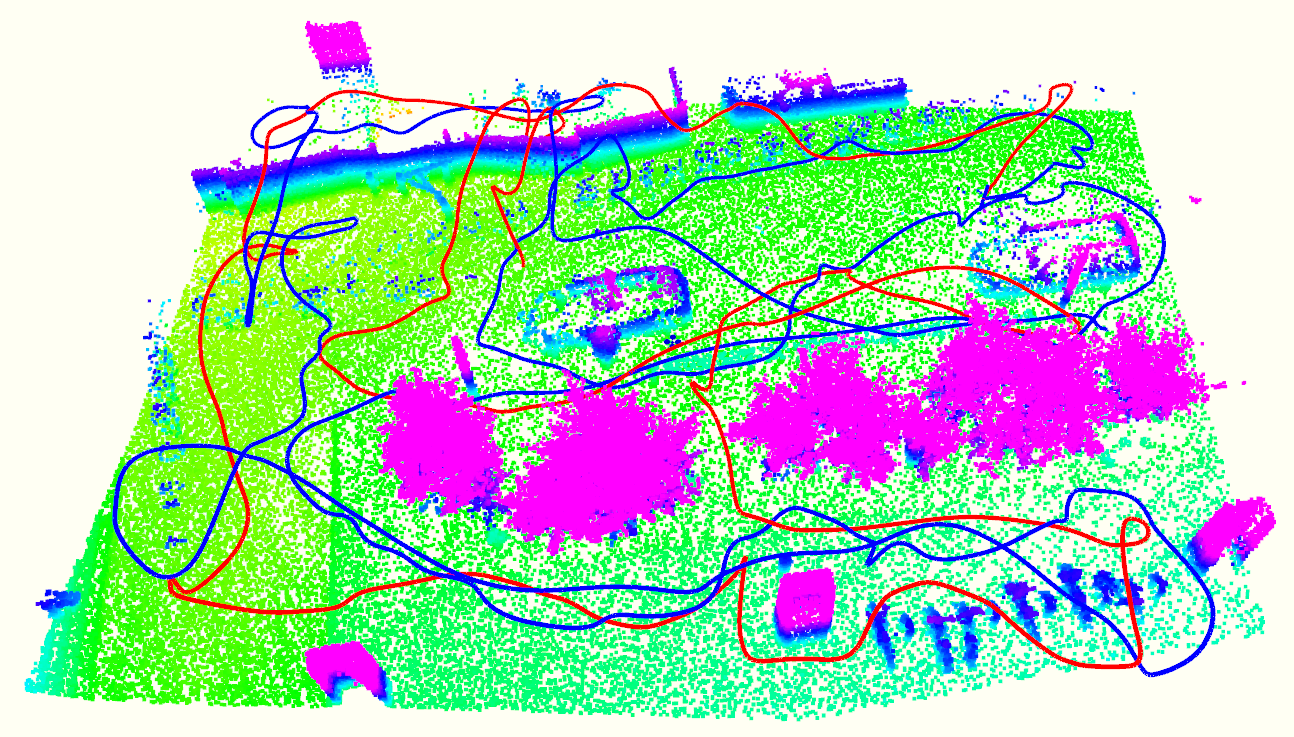}
    \end{minipage}
    }
    \subfigure[]{
    \begin{minipage}[t]{0.31\linewidth}
    \centering
    \includegraphics[width=2.2in]{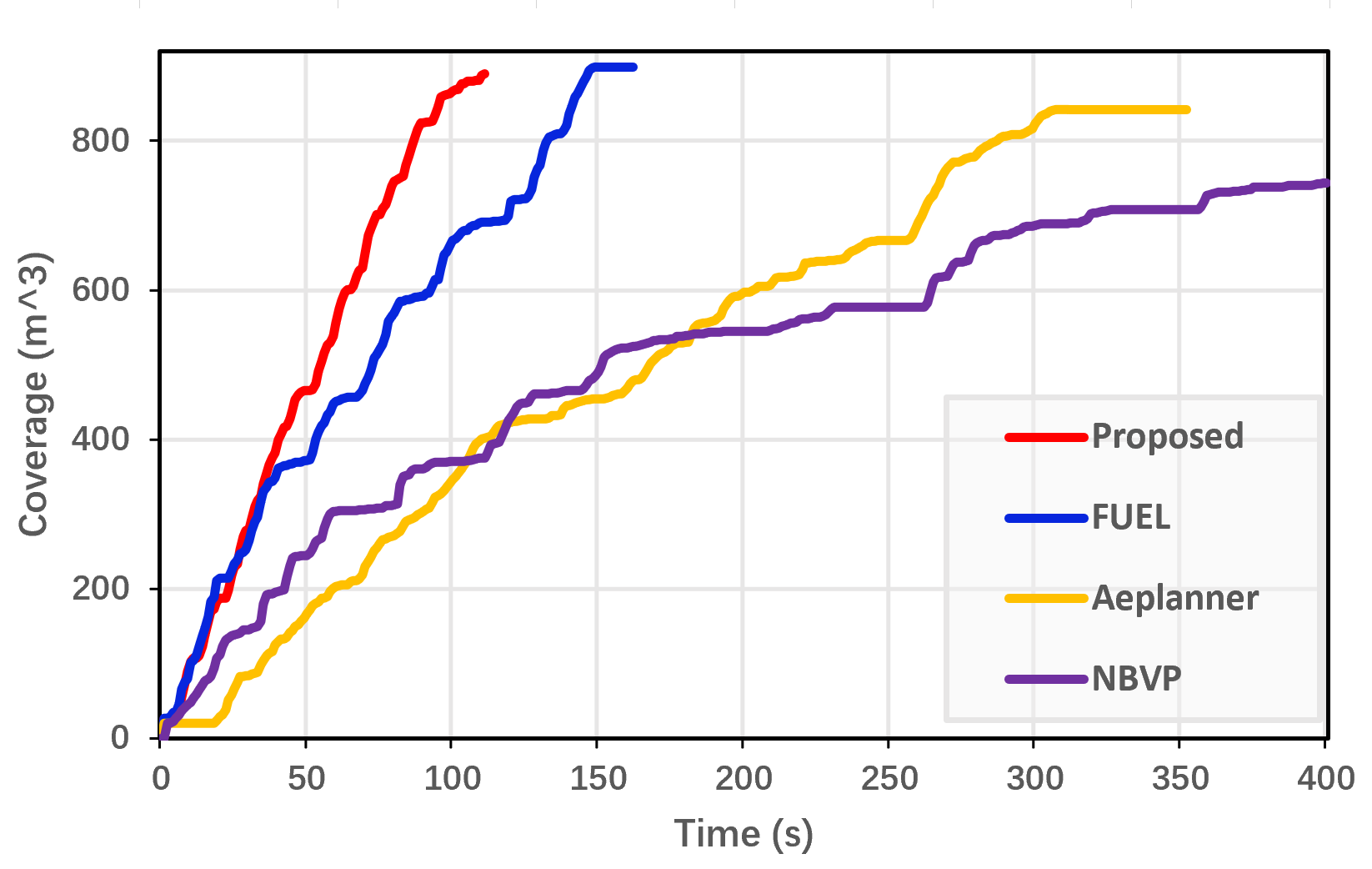}
    \end{minipage}
    }
    \subfigure[]{
    \begin{minipage}[t]{0.31\linewidth}
    \centering
    \includegraphics[width=2.2in]{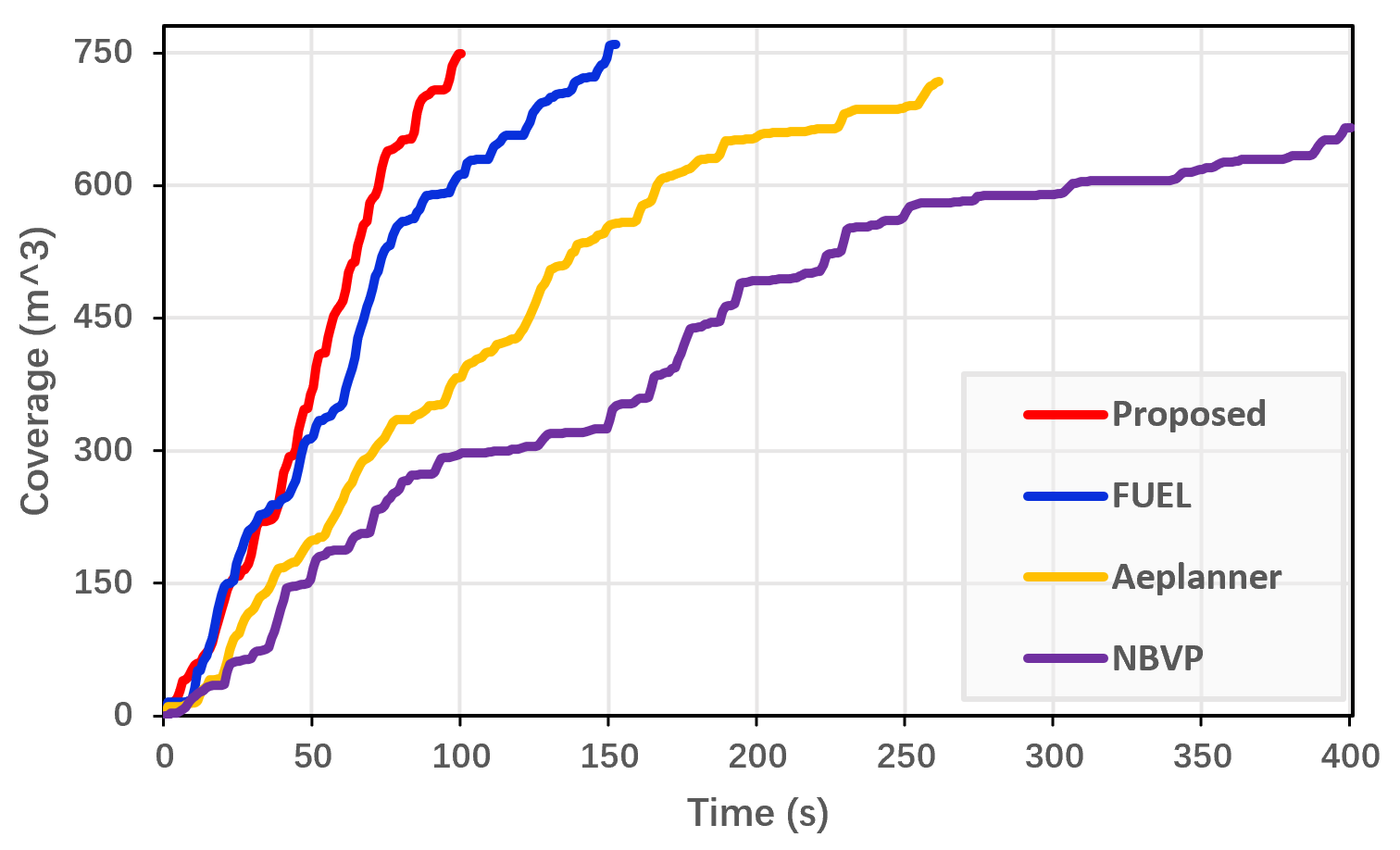}
    \end{minipage}
    }
    \subfigure[]{
    \begin{minipage}[t]{0.31\linewidth}
    \centering
    \includegraphics[width=2.2in]{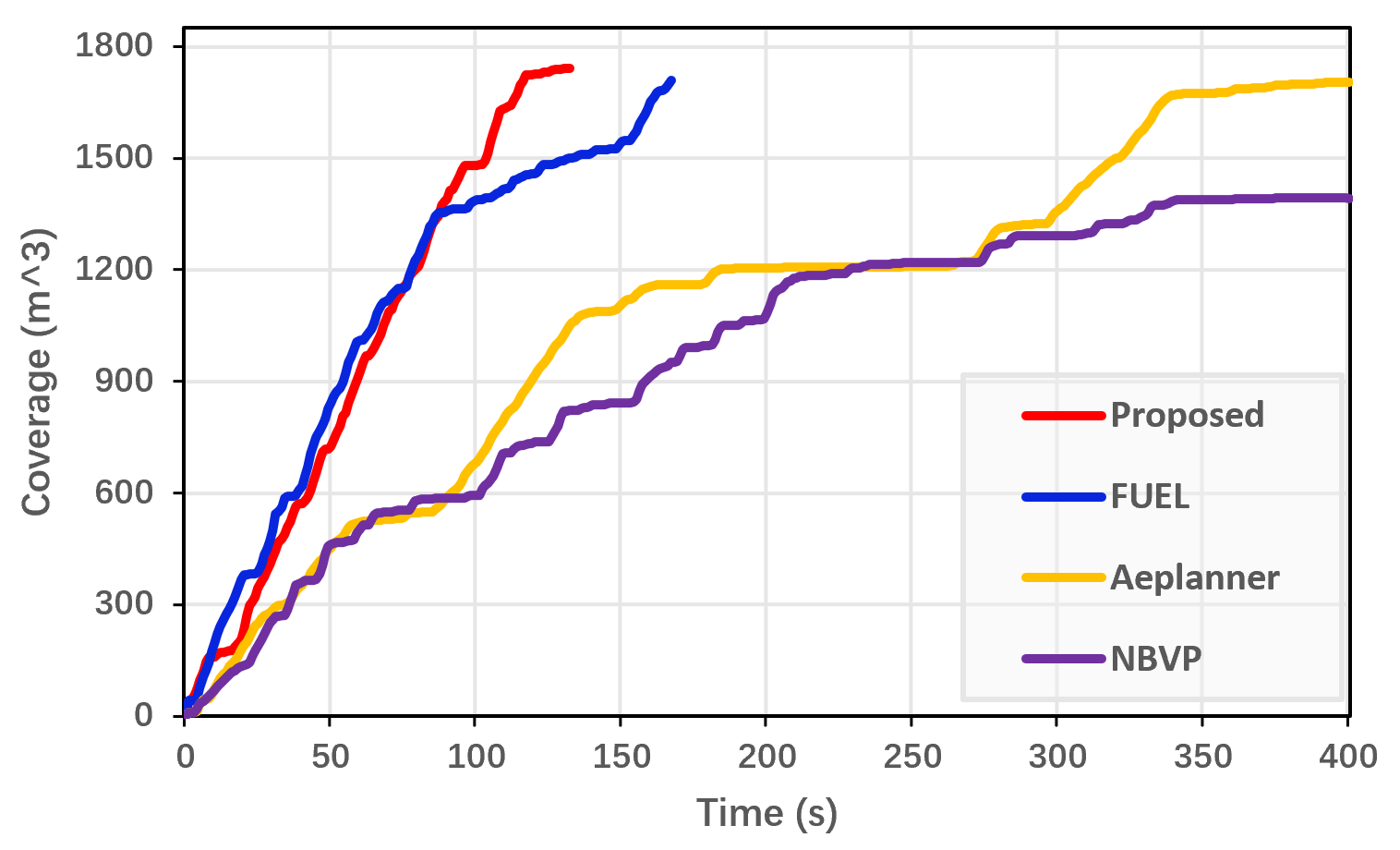}
    \end{minipage}
    }
    \subfigure[]{
    \begin{minipage}[t]{0.31\linewidth}
    \centering
    \includegraphics[width=2.2in]{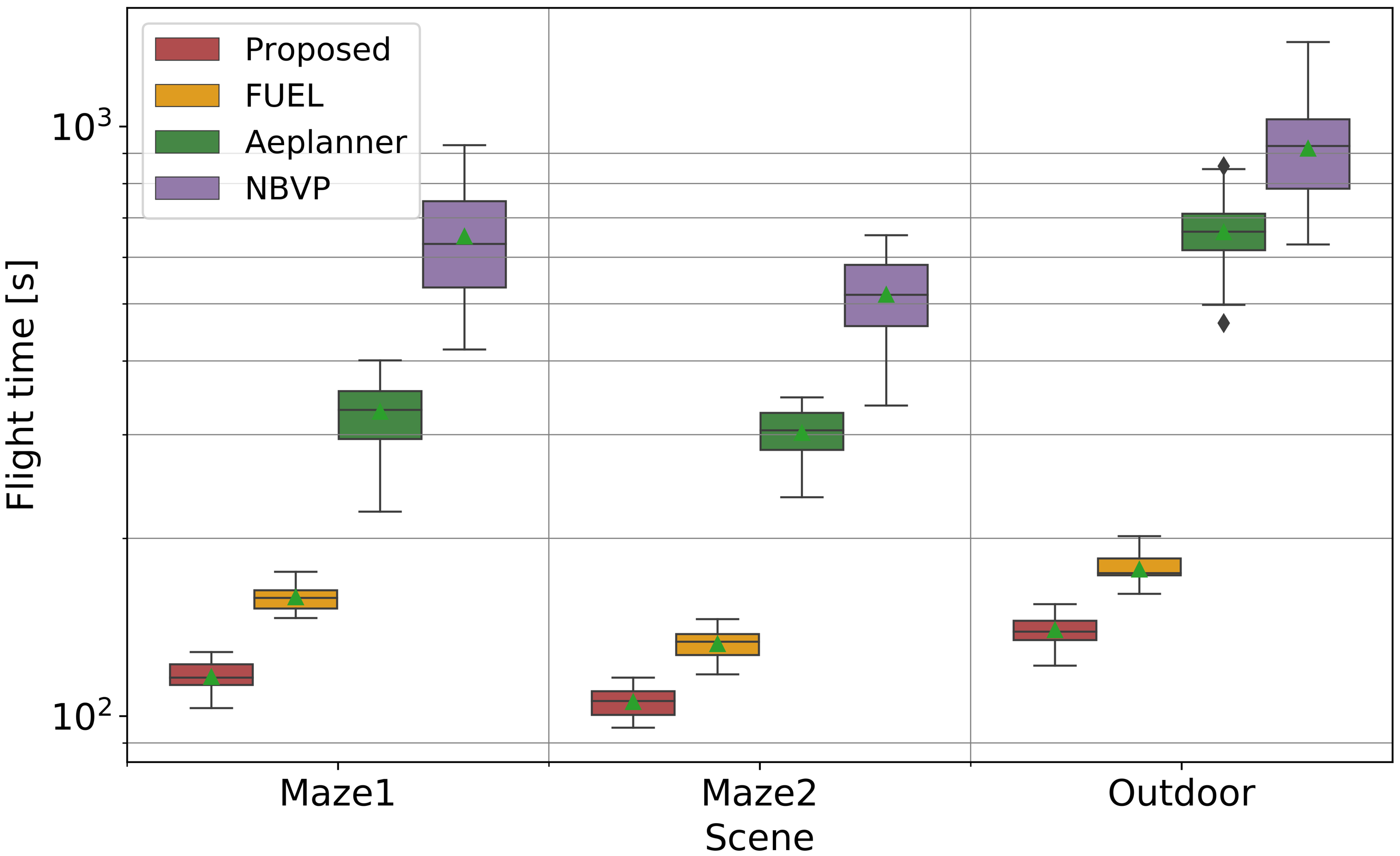}
    \end{minipage}
    }
    \subfigure[]{
    \begin{minipage}[t]{0.31\linewidth}
    \centering
    \includegraphics[width=2.2in]{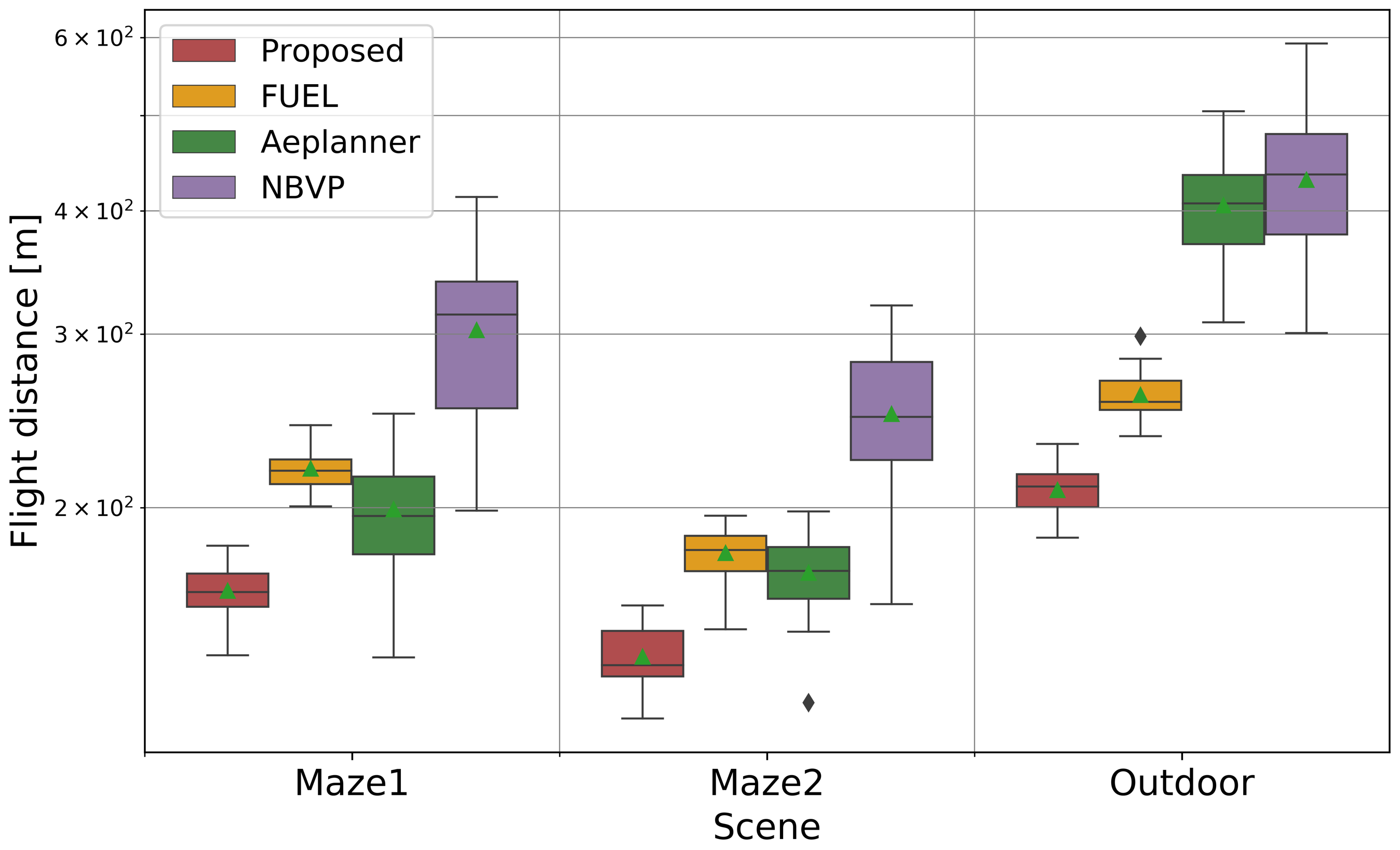}
    \end{minipage}
    }
    \subfigure[]{
    \begin{minipage}[t]{0.31\linewidth}
    \centering
    \includegraphics[width=2.2in]{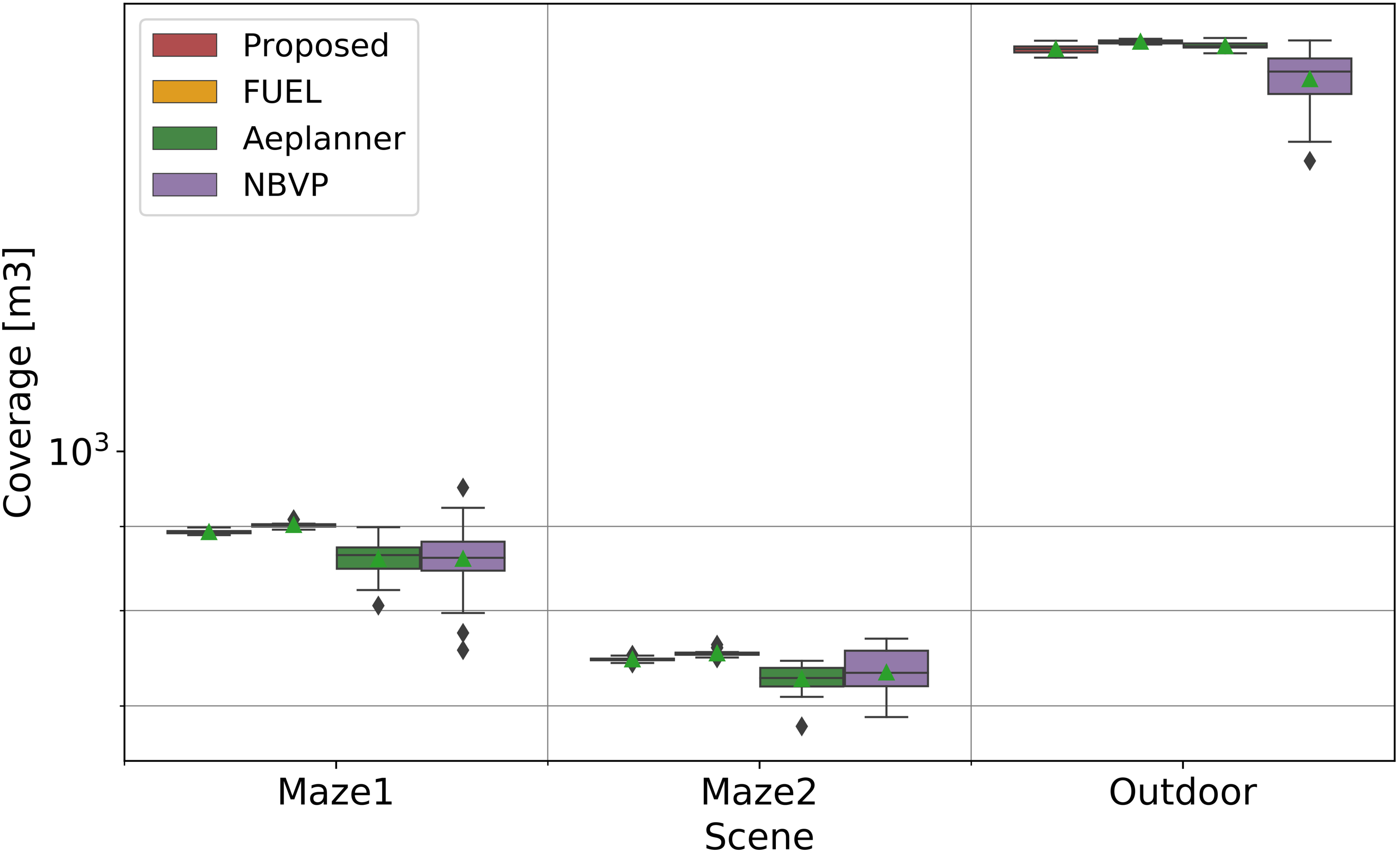}
    \end{minipage}
    }
    \centering
    \caption{Benchmark comparison of the experimental results in different scenarios. (a), (b), and (c) correspond to maze1, maze2, and outdoor scenario respectively (the red and blue trajectories in them represent the best performance of the proposed method and FUEL at the same initial position. And we only show the above two methods for clarity). (d), (e), and (f) respectively illustrate the coverage performance of four comparative experiments in three different scenarios. (g), (h), and (i) respectively show the specific distribution of flight time, flight distance, and coverage in the experimental results (presented in Table. \ref{tab:1}) of the four methods in three different scenarios. More comparison can be found in \url{https://github.com/Zyhlibrary/FAEP}.}
    \label{fig:5}
    \vspace{-6mm}
\end{figure*}

\subsection{Benchmark comparisons}
\begin{table}[]
\caption{PARAMETERS USED FOR SIMULATION}
\label{tab:configuration}
\begin{tabular}{llllll}
\toprule
& Max Velocity & 2.0 m/s       &  & Max Accelerate & 2.0 m/$s^2$      \\
& Max Yaw Rate & 1.0 rad/s   &  & Camera Range   & 4.5 m       \\
& Camera FOV   & [80,60] deg &  & ROS Version    & ROS Melodic \\ \bottomrule
\end{tabular}
\vspace{-5mm}
\end{table}
In the simulation experiment, we compare the proposed method with three state-of-the-art methods in three different environments. The three methods are FUEL \cite{zhou2021fuel}, Aeplanner \cite{selin2019efficient}, and NBVP \cite{bircher2016receding}. For all methods, We adopt their open-source implementation and default configuration (the max nodes of Aeplanner and NBVP is set to 200). And in each scenario, they are tested in 5 different initial positions (5 meters far from each other), and each method is tested 5 times at a position. The configurations we used in the experiments are shown in Table. \ref{tab:configuration}. And we test these methods on a computer with Intel Core i9-9900K@ 3.6GHz, 64GB memory.

\emph{Remark:} Although there are many sensors that can observe wide areas, they are essentially still in the category of limited FOV sensors. Moreover, since we concentrate on the quality of the exploration method, we believe that it would be more convincing that the method could achieve better performance even with the sensor with a narrow observation range, which is more challenging compared with using the ones with a wider observation range.

\subsubsection{Maze Scenario} Maze is the most effective environment to verify the efficiency of exploration method. Therefore, we firstly compare the exploration efficiency in two maze environments: rectangular maze (maze1) and square maze (maze2). The scene range of maze1 is  $30 \times 16 \times 2$ m$^3$, and the maze2 is $20 \times 20 \times 2$ m$^3$. The experimental results are shown in Fig. \ref{fig:5}, and Table. \ref{tab:1}. It can be seen that NBVP takes the longest time and flight distance, and its efficiency is also unstable. Aeplanner is an improved method of NBVP, its efficiency is improved by combining frontier exploration strategy with NBVP. Due to the efficient global coverage path and minimum-time flight path, the proposed method and FUEL have obvious advantages over the above two methods. Not only the flight path is smoother, but also flight time and flight distance are less while ensuring high coverage rate. Meanwhile, due to the fewer back-forth-maneuvers and yaw planning strategy, the proposed method achieves more efficient exploration than FUEL. Compared with FUEL, the exploration time and flight distance of our method are reduced by 26.8\% and 24.8\% respectively in maze1, 20.3\% and 21.5\% in maze2, and the exploration ratio tends to be more linear.

\subsubsection{Outdoor Scenario} We also compare the four methods in nature scenario to verify the stability and efficiency. The scenario contains trees, cars, fences, and other objects, with a range of $20 \times30 \times3$ m$^3$. The experimental results are also shown in Fig. \ref{fig:5}, and Table. \ref{tab:1}. The results show that the exploration time and distance of the four methods increase compared with mazes due to the increase in scene complexity, but the proposed method still has obvious advantages compared with other methods. Compared with NBVP and Aeplanner, our method achieves 5-7 times faster. And compared with FUEL, our method still maintains the advantages of 21.1\% and 20\% in exploration time and flight distance respectively. In addition, we can see that the efficiency of our method and FUEL is basically the same or FUEL achieves better performance in the early stage of exploration, but the advantage of our method gradually opens up in the middle and late stages, this is because the high information gain will be obtained regardless of which frontier is explored in priority in the early stage, so the quality of the global exploration sequence does not affect the exploration efficiency of the early stage. And since our method will promptly explore the frontiers that may have low information gain but will cause the back-and-forth maneuvers, the efficiency of our method may not be the most efficient in the early stages. However, as exploration advances, back-and-forth maneuvers caused by the low-quality global exploration sequence gradually emerge, which leads to the reduced efficiency of FUEL. And due to the high-quality exploration sequence, as shown in Fig. \ref{scene1}\ref{scene2}\ref{scene3}, our method has significantly fewer back-and-forth maneuvers, so our method achieves better performance in the total exploration effectiveness.
\begin{figure*}[!ht]
    \subfigure[]{
    \begin{minipage}[t]{0.31\linewidth}
    \centering
    \includegraphics[width=2.0in, height=1.3in]{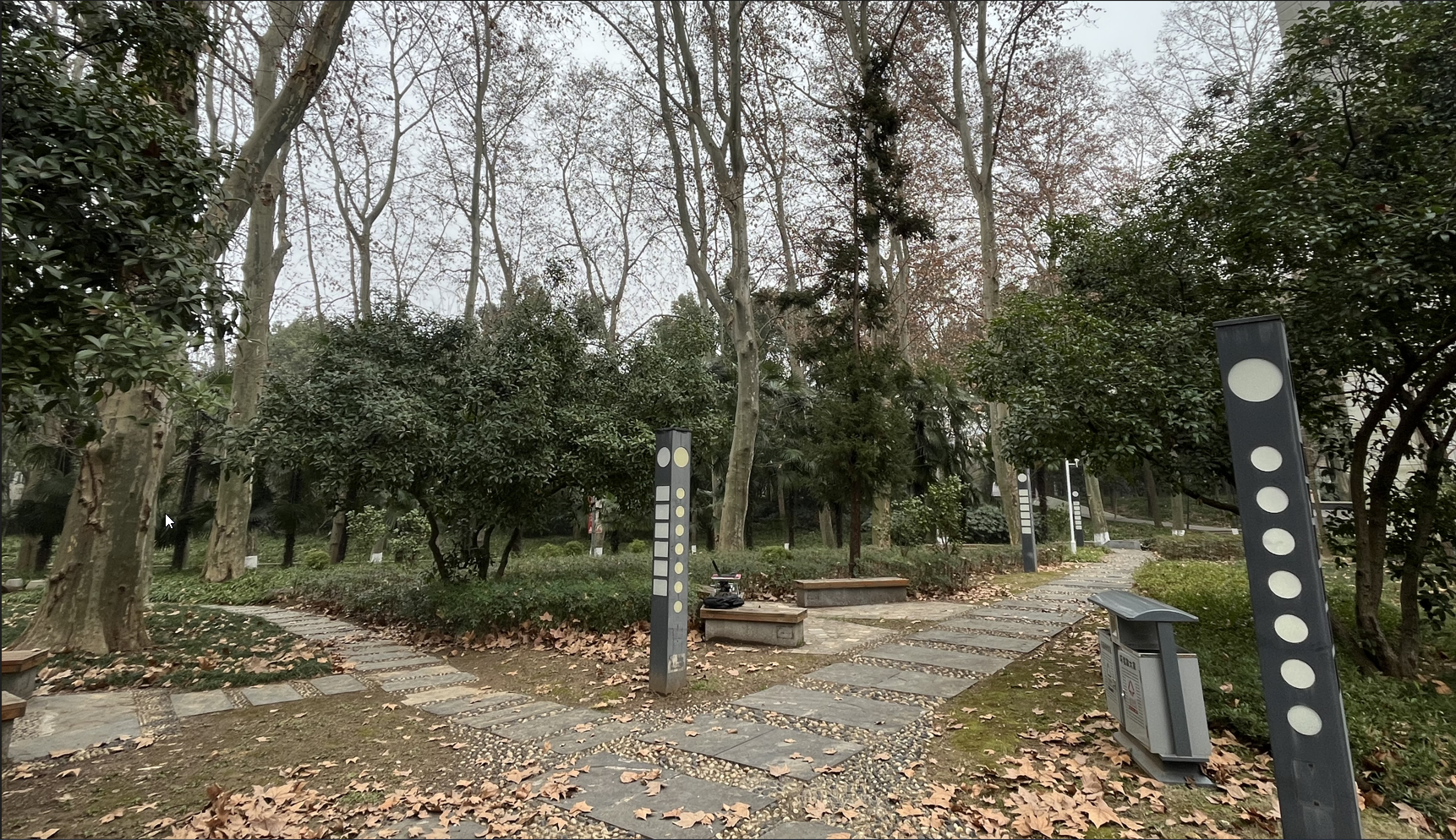}
    \end{minipage}
    }
    \subfigure[]{
    \begin{minipage}[t]{0.31\linewidth}
    \centering
    \includegraphics[width=1.9in]{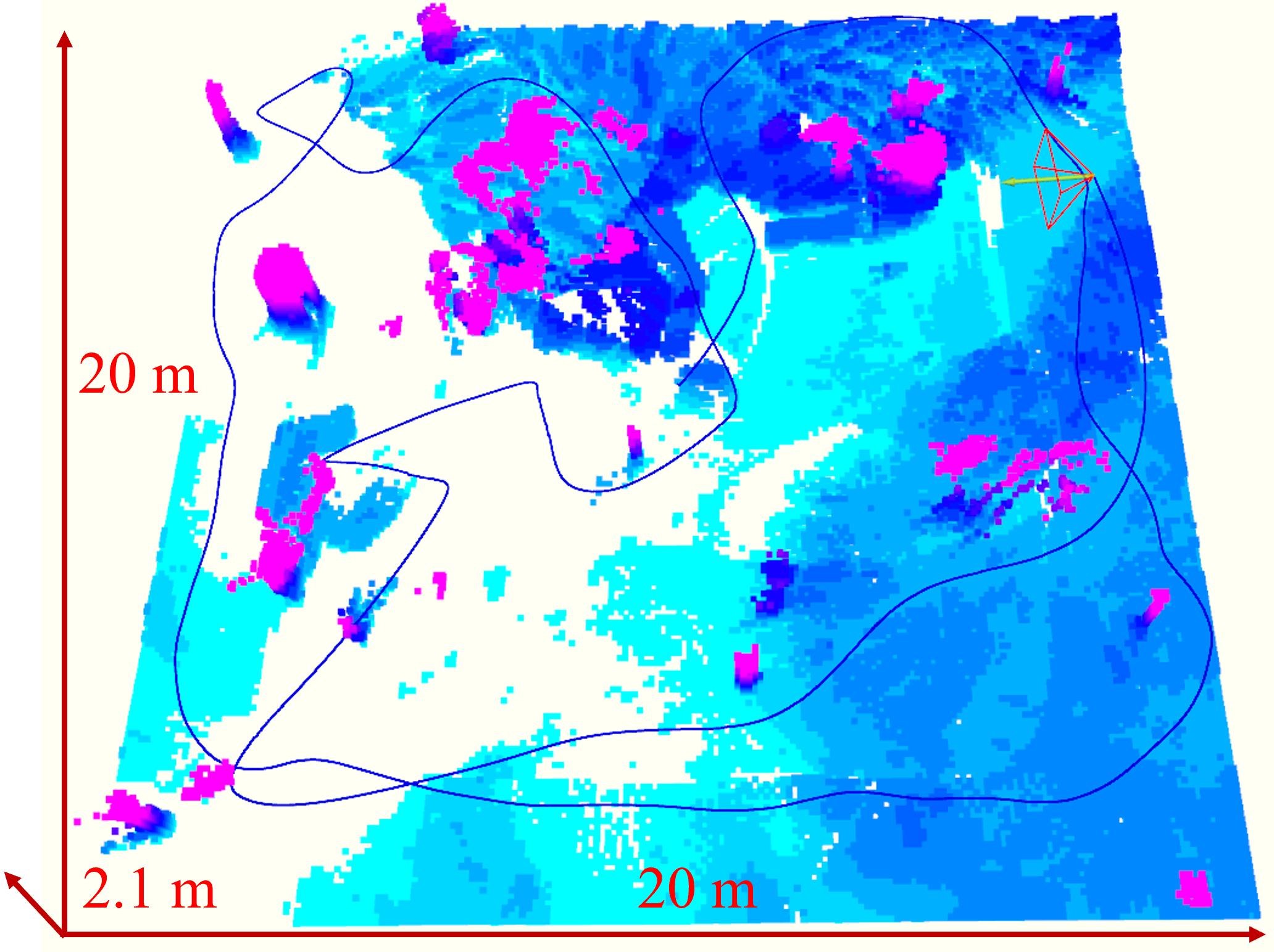}
    \end{minipage}
    }
    \subfigure[]{
    \begin{minipage}[t]{0.31\linewidth}
    \centering
    \includegraphics[width=2.2in]{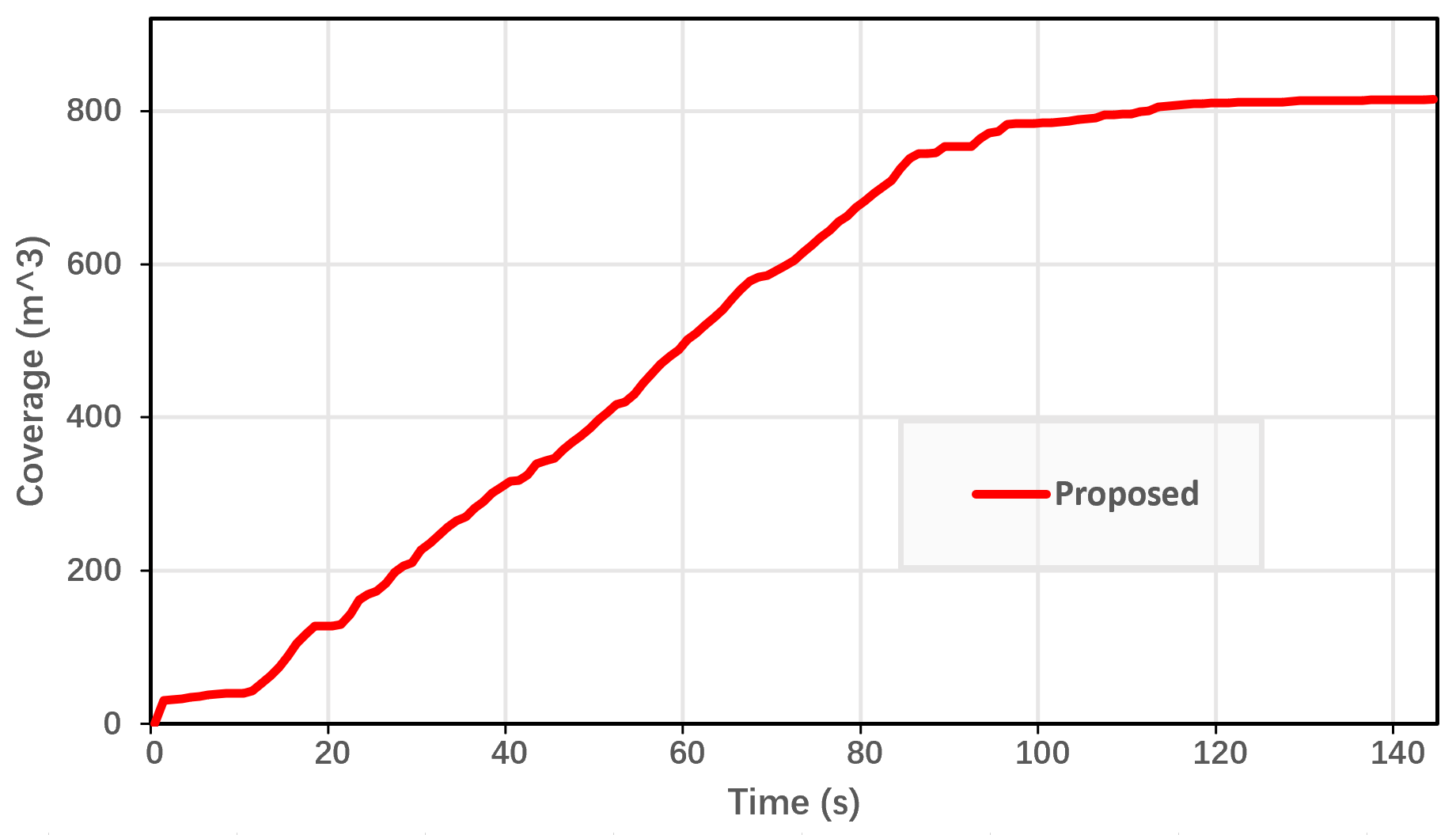}
    \end{minipage}
    }
    \subfigure[]{
    \begin{minipage}[t]{0.31\linewidth}
    \centering
    \includegraphics[width=2.0in, height=1.3in]{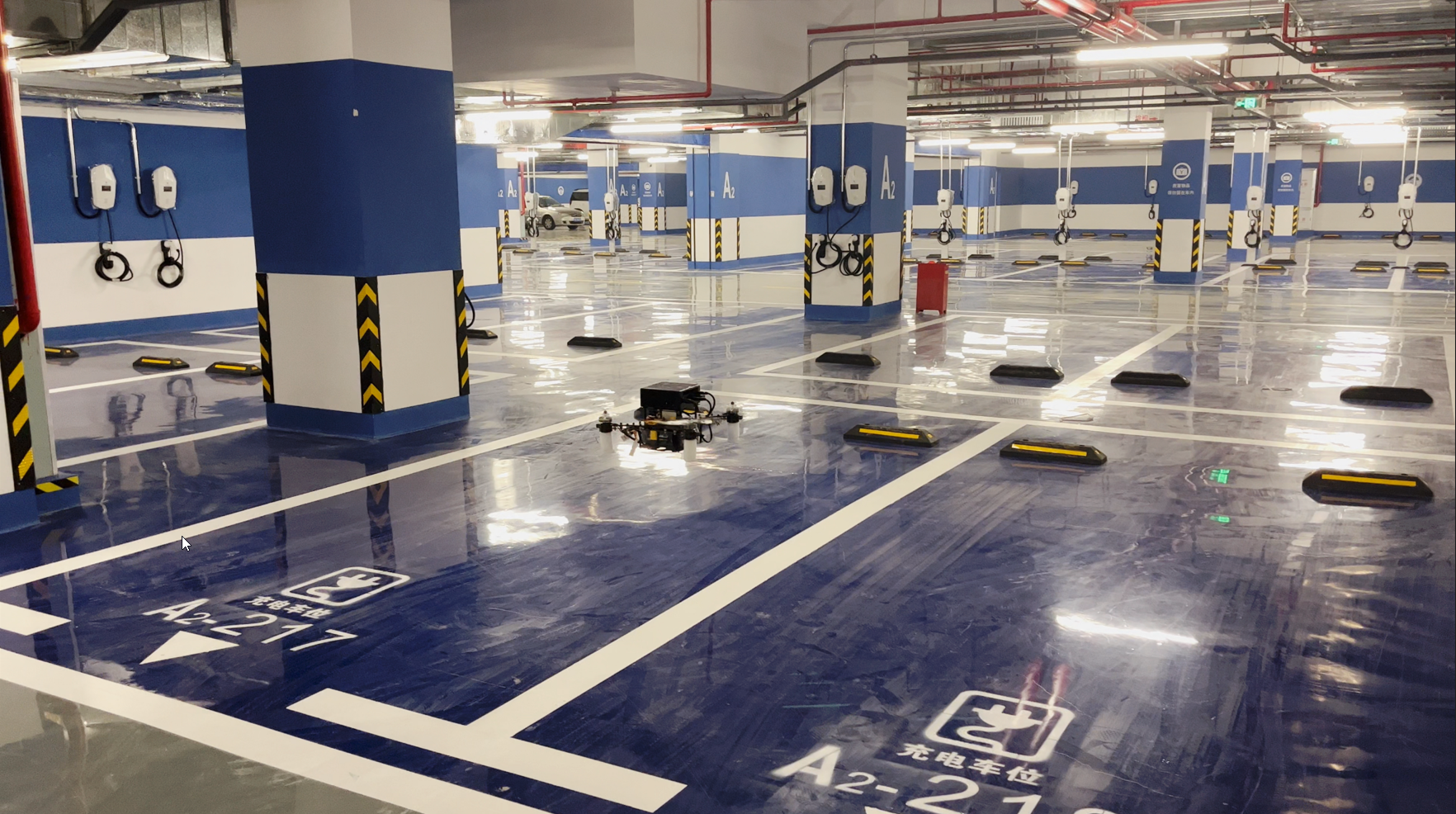}
    \end{minipage}
    }
    \subfigure[]{
    \begin{minipage}[t]{0.31\linewidth}
    \centering
    \includegraphics[width=1.9in]{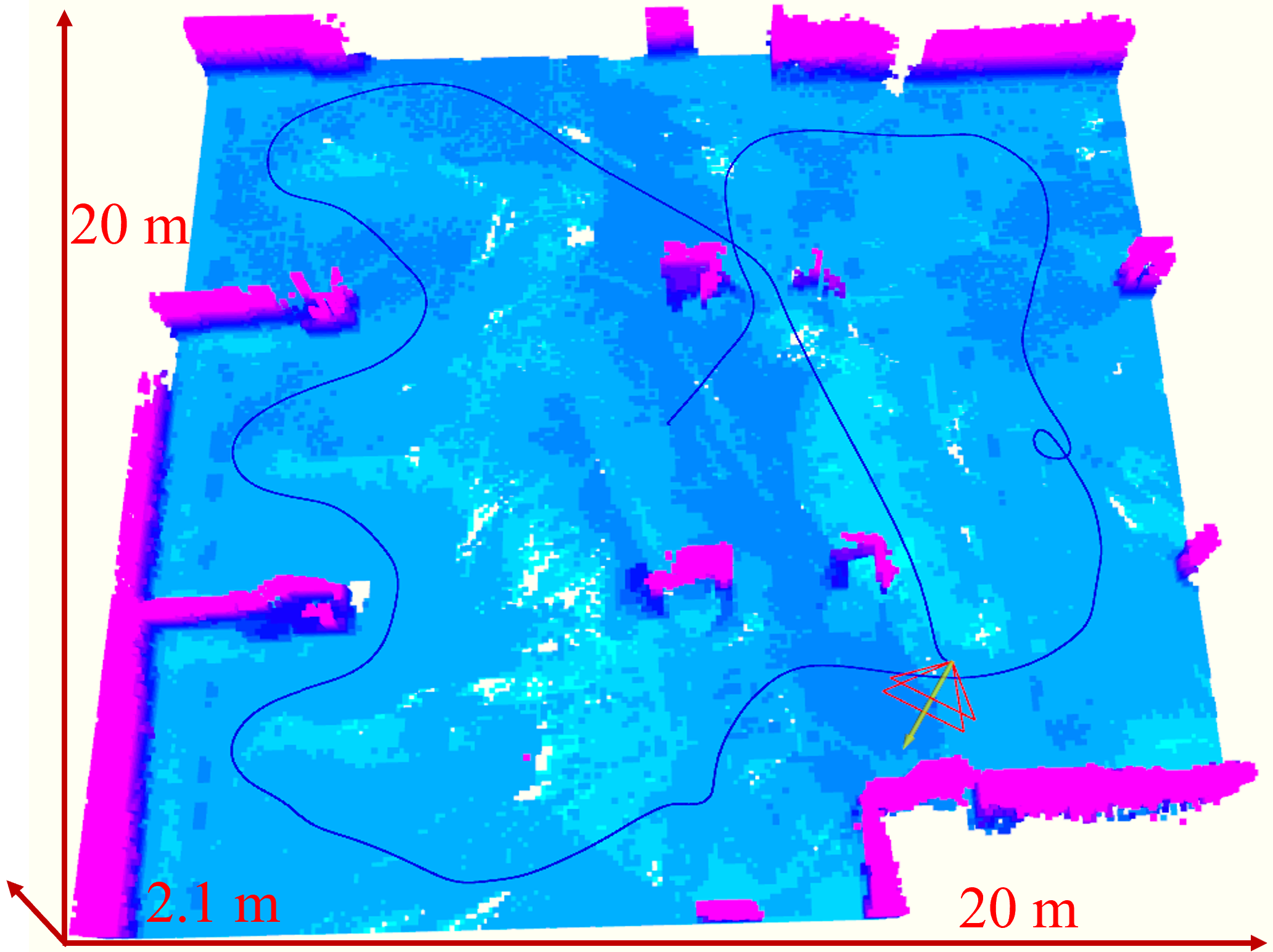}
    \end{minipage}
    }
    \subfigure[]{
    \begin{minipage}[t]{0.31\linewidth}
    \centering
    \includegraphics[width=2.2in]{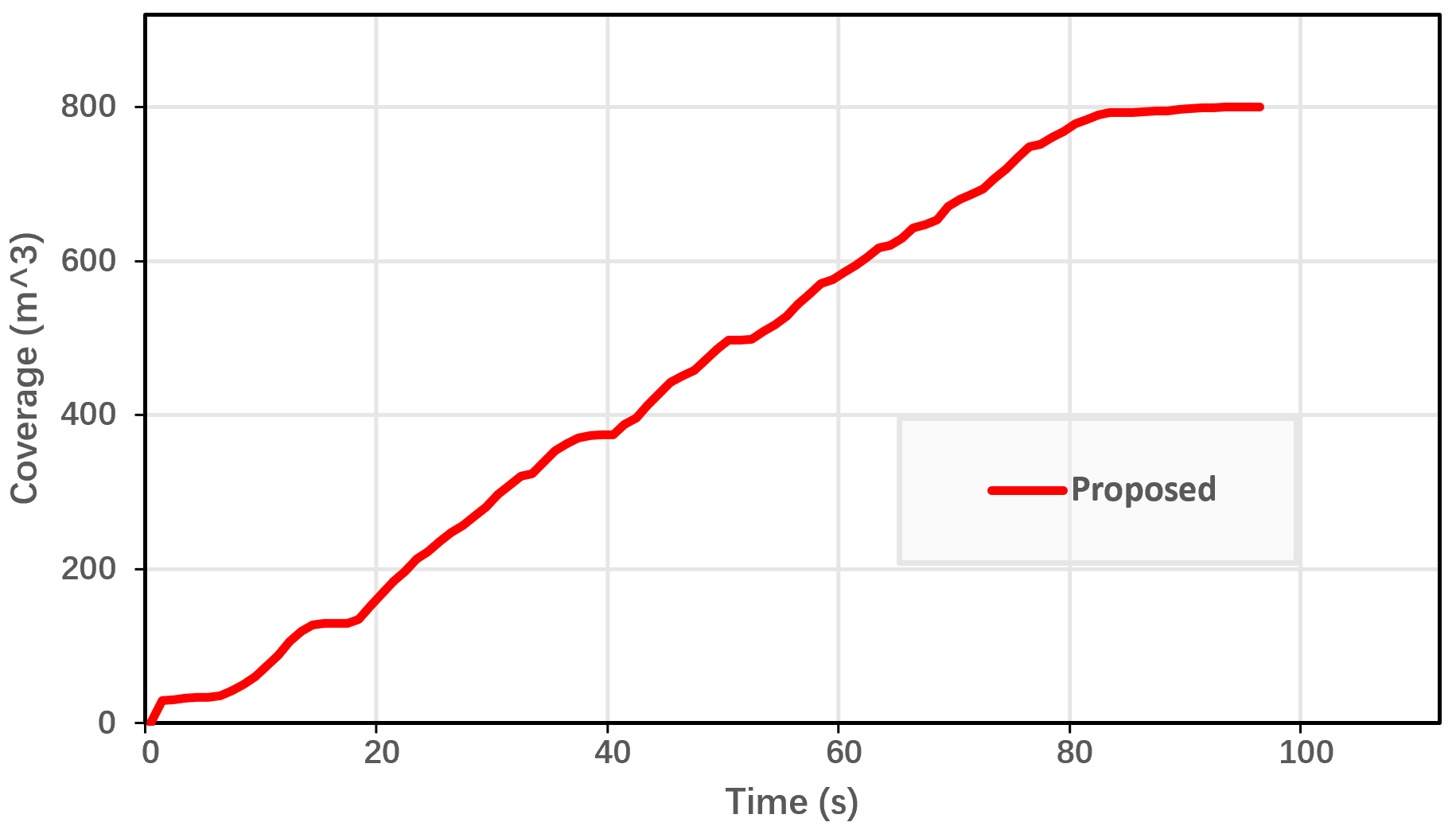}
    \end{minipage}
    }
    \centering
    \caption{The results of real-world experiments. (a), (b), and (c) are the experiment results in the park. (d), (e), and (f) are the results in the underground parking lot. It should be noted that the paths in (b) and (e) are the actual exploration trajectory, which are obtained by recording the position of UAV provided by SLAM. (e) and (f) show the change in exploration coverage relative to the exploration time, which can be obtained by the number of known voxel grids. Videos of the experiments can be found at \url{https://youtu.be/0Y671mEwJ\_A}}.
    \label{fig:real-world}
    \vspace{-5mm}
\end{figure*}
% \usepackage{multirow}
\begin{table}[!t]
    \begin{center}
		\caption{AVERAGE COMPUTATION TIME IN DIFFERENT SCENES}
		\label{tab:2}
		\centering
		\setlength{\tabcolsep}{4mm}{
			\renewcommand{\arraystretch}{1.1} {% Default value: 1
\begin{tabular}{ccccc}
\toprule

\multirow{2}{*}{\textbf{Method}} & \multicolumn{4}{c}{\textbf{Experimental results in three scenes (ms)}}                                            \\ \cline{2-5} 
                                 & \multicolumn{1}{l}{Maze1} & \multicolumn{1}{l}{Maze2} & \multicolumn{1}{l}{Outdoor} & \multicolumn{1}{l}{Avergae} \\ 
Aeplanner                        & 623                       & 756                       & 761                         & 713                         \\
NBVP                             & 794                       & 912                       & 926                         & 877                         \\
FUEL                             & 29                        & 25                        & 33                          & 29                          \\
Proposed                         & 24                        & 21                        & 31                          & 25                          \\ \bottomrule 
\end{tabular}
}}
	\end{center}
	\vspace{-5mm}
\end{table}
\begin{figure}[!t]
	\vspace{0mm}
	\centering
	\includegraphics[width=.8\linewidth]{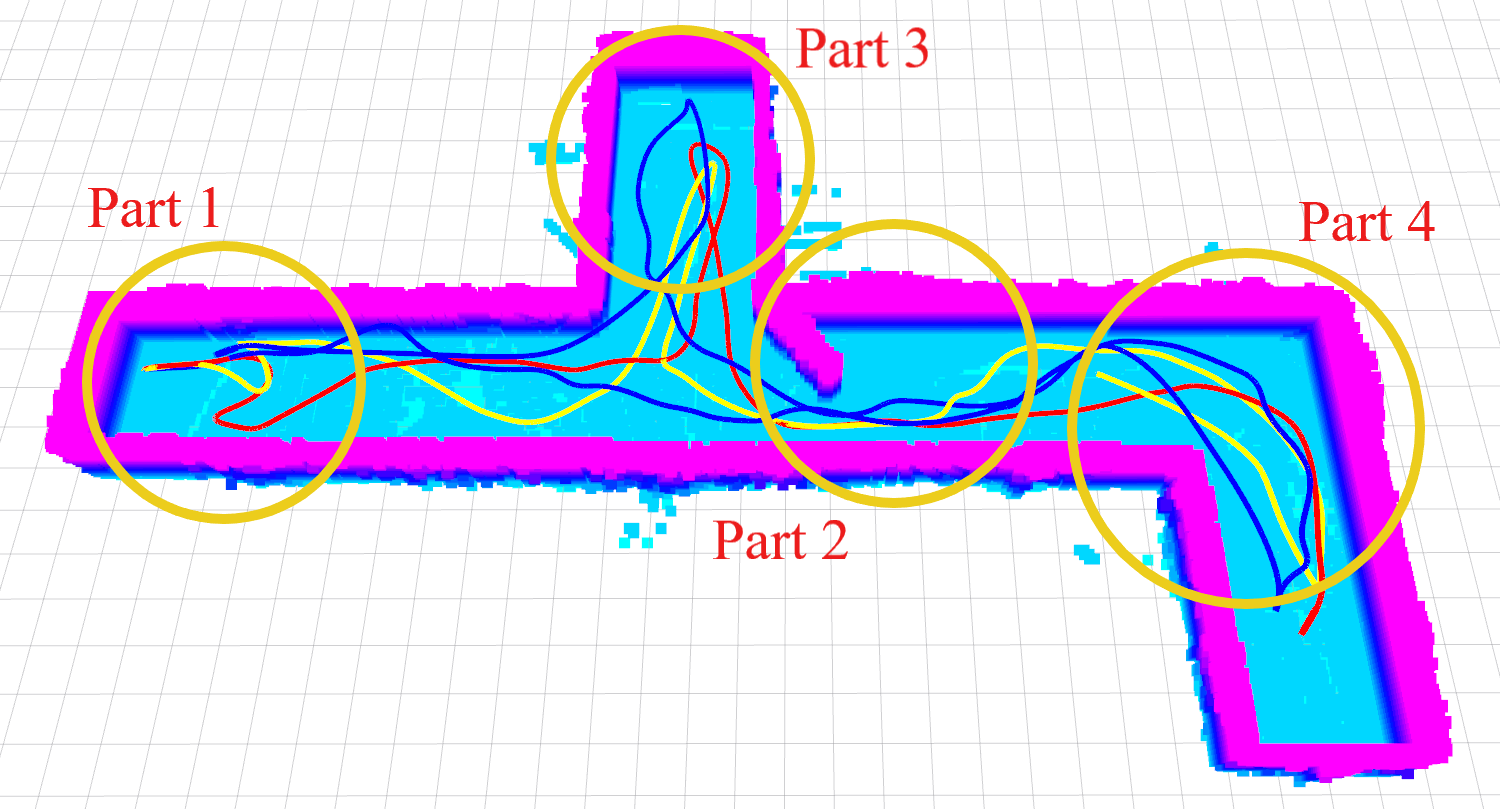}
	\caption{A corridor scenario that is used to clearly explain the enhancement of our method.Red trajectory and blue path are actual exploration trajectory of our method and FUEL respectively. And the yellow trajectory is the exploration trajectory of our method without adaptive yaw planning part. Orange circles represent the key parts that cause the efficiency differences.}
	\label{fig:add2}
	\vspace{-5mm}
\end{figure}\textbf{}
\subsubsection{Further Evaluation} We can also see that the fluctuation range of the results generated by our method in different initial positions is very limited, which implies that different initial positions have a limited impact on the final exploration performance of our method. This is because once the exploration area is determined, no matter where the initial position is, it only affects the exploration area and exploration trajectory in the early stage, and our method will drive the UAV to explore the area according to our exploration principles. In addition, we have conducted a comparison of computation time for the proposed method and the three state-of-the-art methods, the comparison result is shown in Table. \ref{tab:2}. We can see that due to the use of a heuristic framework that contains a frontier information structure, our method and FUEL can easily maintain crucial information in the entire space. Therefore, our method and FUEL achieve better performance in computation time.
\begin{figure}[!ht]
    \subfigure[]{
    \begin{minipage}[t]{0.45\linewidth}
    \centering
    \includegraphics[width=1.8in]{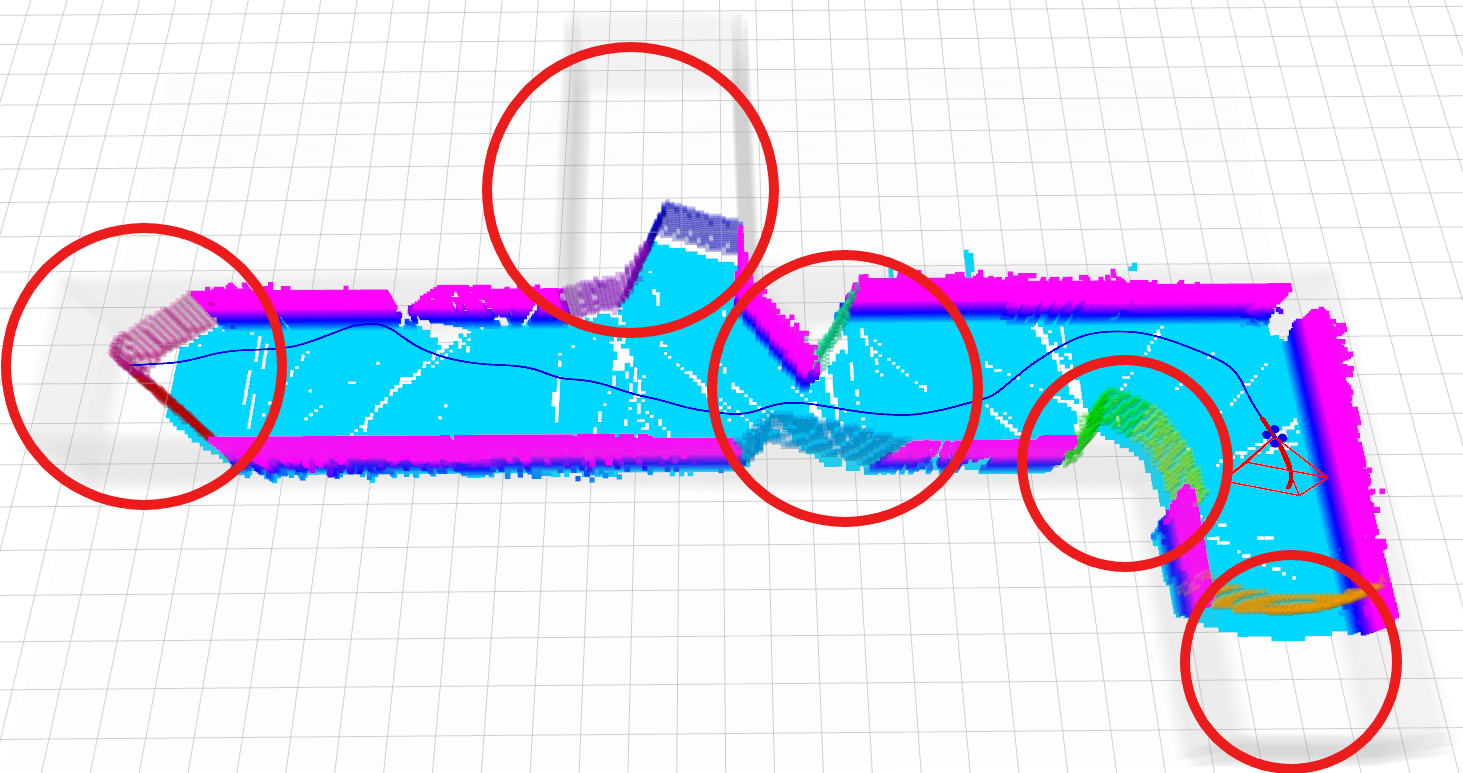}
    \end{minipage}
    }
    \subfigure[]{
    \begin{minipage}[t]{0.45\linewidth}
    \centering
    \includegraphics[width=1.8in]{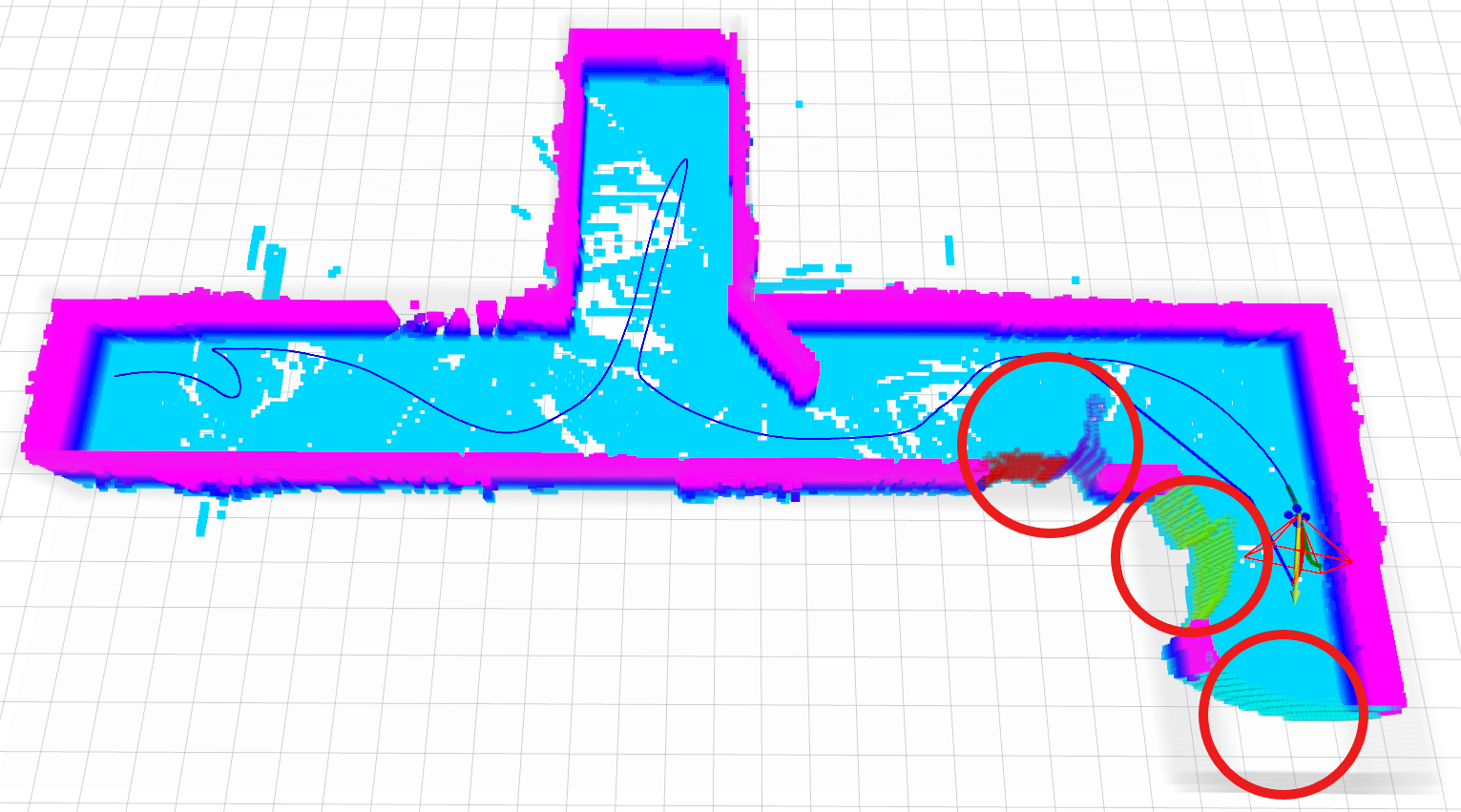}
    \end{minipage}
    }
    \subfigure[]{
    \begin{minipage}[t]{0.45\linewidth}
    \centering
    \includegraphics[width=1.8in]{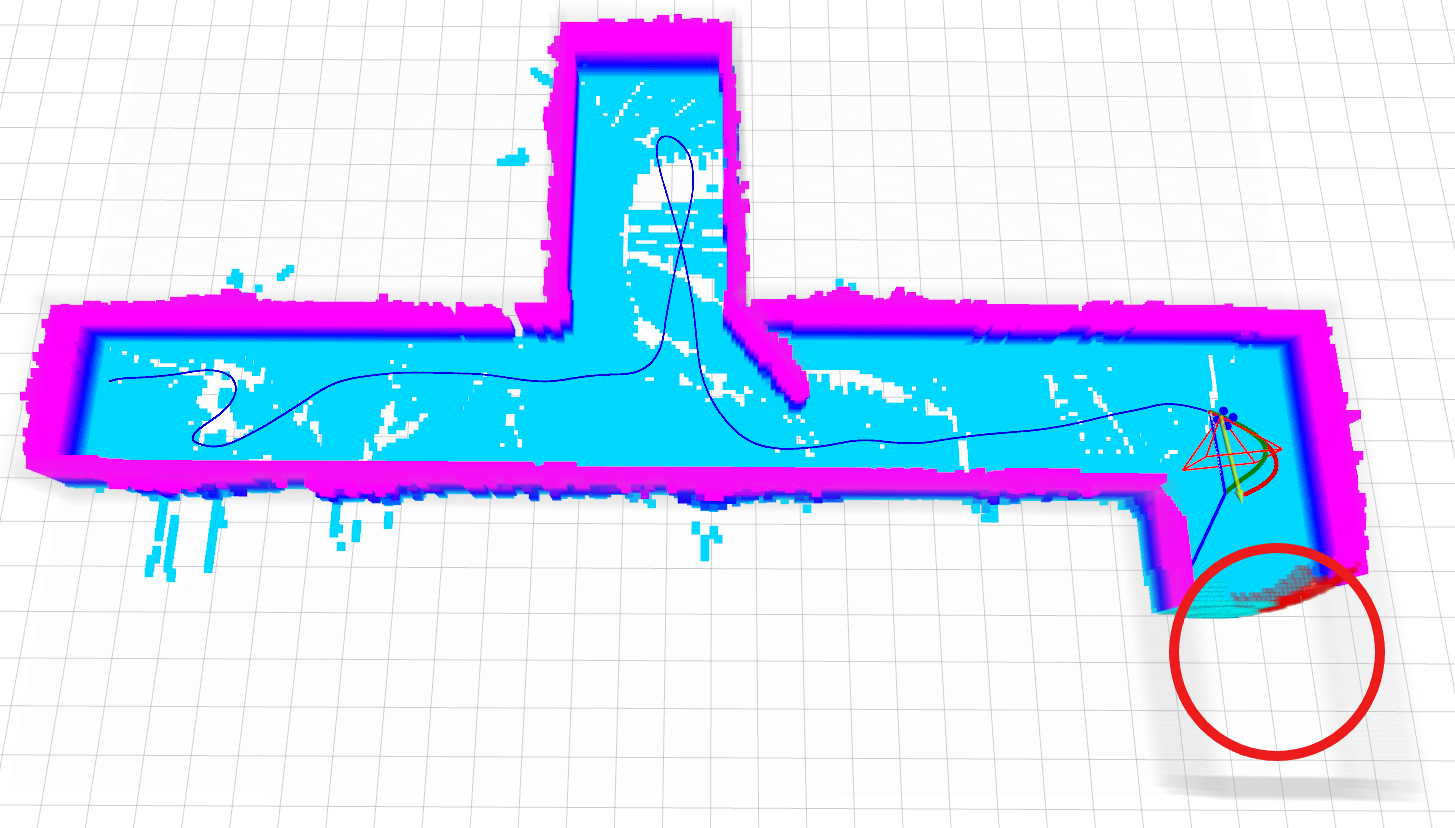}
    \end{minipage}
    }
    \centering
    \caption{The global exploration status when UAV flies to the Part 4. (a), (b), and (c) are the results of FUEL, Method-1, and our method respectively. The red circles indicate areas that have not been explored.}
    \label{fig:exploration_status}
    \vspace{-6mm}
\end{figure}

In addition, to cover our contribution more clearly, as shown in Fig. \ref{fig:add2}, we design a simple yet effective comparison experiment in a corridor for our method, our method without adaptive yaw planning (named by Method-1 for clarity), and FUEL. To finish the exploration, FUEL takes 45.4 s and 74.4 m, Method-1 takes 32.2 s and 60.5 m. And our method achieves the best performance with 30.7 s and 55.9 m. We can see from Fig. \ref{fig:add2} that our trajectory is more reasonable and has fewer back-and-forth maneuvers, which is mainly related to four parts: Part 1, Part 2, Part 3, and Part 4. Part 1 is the area where the exploration begins. Due to the limitation of the FOV, some small areas will be left behind. Part 2 contains a corner that will cause an area that is difficult to explore promptly. Part 3 is a protruding corridor. Part 4 is an area containing a large corner that is easy to leave areas when turning. For Part 1 and Part 2, since we prioritize the exploration of small independent areas mentioned in Sect. \ref{thirda}, our method and Method-1 will explore the areas promptly even with a longer stay in these areas, which slightly sacrifices the local efficiency but avoids the subsequent larger impact caused by back-and-forth maneuvers. Therefore, we can see that the trajectories of our method and Method-1 have a clear detour to cover the area in the two parts. For Part 3, due to the use of the spiral filling concept as shown in Fig. \ref{fig:2}(a), our method (including Method-1) will explore the part in priority to reduce inefficient paths. In Part 4, since our method uses the adaptive yaw planning mentioned in Sect. \ref{thirdb}, we will cover more frontiers by yaw change. Therefore, compared with Method-1 and FUEL, we will not leave unexplored areas that need to go back and explore again in Part 4. For proving the above analysis, as shown in Fig. \ref{fig:exploration_status}, we also provide the global exploration statuses of the three methods when UAV flies to the Part 4, the distribution of unexplored areas is consistent with our analysis.
\vspace{-5mm}
\subsection{Complexity Analyzing}
Our method has three main parts: global path solving by ATSP, local path planning and yaw planning, and map maintenance. In the first part, we use LKH\cite{helsgaun2000effective} to solve the ATSP in global planning. And as proved in \cite{helsgaun2000effective}, the complexity of LKH is $O(n^2)$. Thus, our computation complexity in this part is $O(N_{cls}^2)$, where $N_{cls}$ represents the number of the frontier and its size depends on the length of the boundary between known and unknown areas. Once the length of the boundary exceeds the distance threshold (about 2m), the boundary will be split into two frontiers. And following this principle, $N_{cls}$ can be determined. In planning part, we use the B-spline to generate the local trajectory and yaw path, so the computational complexity of this part is $O(N_c)$, where $N_c$ is the number of the control points of B-spline and $N_c$ is approximately equal to the value of the planning horizon divided by the distance interval of adjacent control points (about 0.5m). In map maintenance, since we only update the local voxel grids, the time complexity is $O(N_v)$, where $N_v$ is the number of voxel grids affected by the data of the sensor, and its size is determined by the FOV of the sensor.
\vspace{-2mm}
\subsection{Real-world Experiments}
\begin{figure}[t]
	\centering
	\vspace{2mm}
	\includegraphics[width=.7\linewidth]{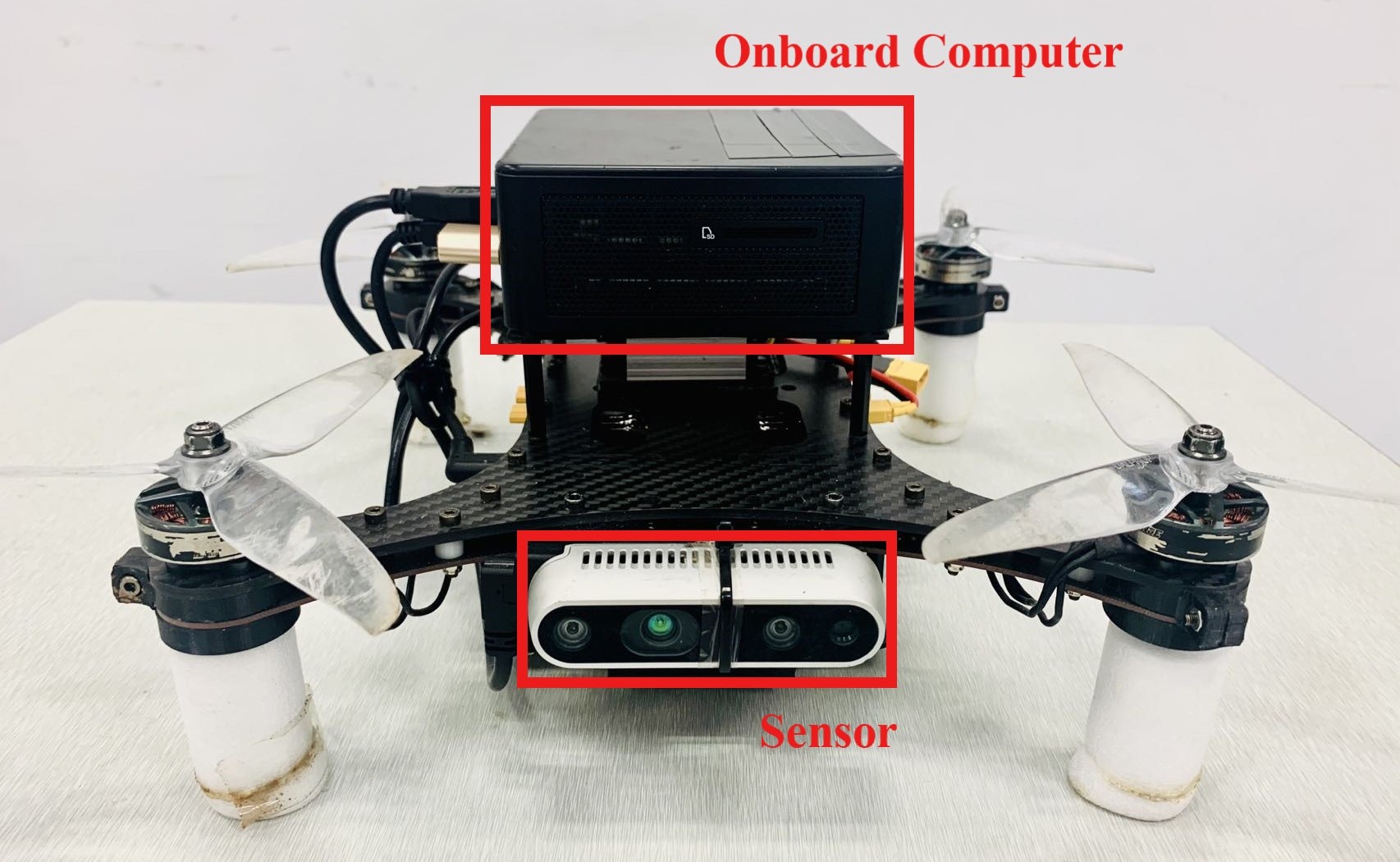}
	\caption{The customized quadrotor used in real-world experiments.}
	\label{fig:uav}
	\vspace{-3mm}
\end{figure}\textbf{}
\begin{table}[!t]
\caption{SYSTEM PARAMETERS OF THE CUSTOMIZED QUADROTOR}
\label{tab:UAV}
\begin{tabular}{lllll}
\toprule
Wheelbase & 330 mm       & Flight Control & CUAV V5+      \\
Motor  & TMOTOR F90 KV1300   & Propeller Size   & 7 inch       \\
Sensor   & Intel RealSense D435 & Computer    & Intel NUC \\ \bottomrule
\end{tabular}
\vspace{-5mm}
\end{table}
In order to verify the effectiveness of the proposed method, we also conduct two real-world experiments in the park and underground parking lot by using our customized quadrotor, as shown in Fig. \ref{fig:uav}. We equipped our UAV with a limited FOV sensor, and use\cite{VINS} to provide the quadrotor state. All the modules run on an onboard computer NUC equipped with Intel Core i5-1135G7@ 2.40GHz, 16GB memory and ROS Melodic. The detailed system parameters are shown in Table. \ref{tab:UAV}. In the experiments, we set dynamic limits as $v_{max} = 1.0$ m/s, $a_{max} = 1.0$ m/s$^2$ and $\dot \xi_{max} = 1.0$ rad/s.

At first, to validate our method in natural scenarios, we conduct an exploration experiment in park. The scenario contains trees, bushes, stone stools, and other objects. We bound the scenario for exploration by a $20 \times20 \times2.1$ m$^3$ box. The exploration results are shown in subgraph (a), (b) and (c) of Fig. \ref{fig:real-world}. The exploration time of the whole process is 144.5 s, and the flight distance is 131.0 m. And it should be noted that the minimum height of the exploration area pre-set by us is $-$0.1 m, but the park is an uneven area that contains a large low-lying area, which causes the visually blank area (no point clouds) in subgraph (b) of Fig. \ref{fig:real-world}. And then, to verify our method in the underground scenario, we also conduct exploration experiments in an underground parking lot, which is bounded by a $20 \times 20 \times2.1$ m$^3$ box. The experiment results are shown in subgraph (d), (e), and (f) of Fig. \ref{fig:real-world}. The exploration time and flight distance of the whole exploration process are 94.3 s and 90.2 m respectively. In addition, the data collected in the two real-world experiments are stored in an occupancy grid map with 0.1 m resolution (subgraph (b) and (e) of Fig. \ref{fig:real-world}), and the size of the two maps are 1.82 MB and 3.13 MB respectively. The above two experiments prove that our method can achieve the exploration task effectively and safely by using the limited FOV sensor in natural experiments and indoor environments. We also provide a video demonstration in Fig. \ref{fig:real-world} for readers to get more details.
\section{Conclusion And Future Work}
This paper proposes an autonomous exploration planner for fast unknown environment mapping by UAV equipped with limited FOV sensors. Firstly, we design 
a comprehensive frontiers exploration sequence generation method, which not only considers the cost of flight-level (distance, yaw change, and velocity direction change) but also considers the spatial features of the frontier to reduce the back-and-forth maneuvers. Secondly, according to the distribution of frontiers and flight state of UAV, an adaptive yaw planning method is proposed to cover more frontiers when the time required for two-stage yaw change is less than or approximately equal to the flight time of the current trajectory. Finally, path planning and dynamic replanning is adopted to increase the stability and fluency of the flight process by selecting the dynamic start point and corresponding replanning strategy. Both simulation and real-world experiments verify the efficiency of our method.

% References

\bibliographystyle{Bibliography/IEEEtranTIE}
\bibliography{Bibliography/IEEEabrv,Bibliography/BIB_xx-TIE-xxxx}\ %IEEEabrv instead of IEEEfull

%\vspace{-1cm}

\end{document}